\documentclass[10pt,journal,compsoc]{IEEEtran}

\usepackage{booktabs} 
\usepackage{url}
\usepackage{amsmath}
\usepackage{float}
\usepackage{subfig}
\usepackage{graphicx}
\usepackage{color}
\usepackage[noend]{algorithmic}
\usepackage{algorithm}
\usepackage{multirow}
\usepackage{verbatim}
\usepackage{amsmath}
\usepackage{amsthm}
\usepackage{indentfirst}
\usepackage{breqn}
\usepackage{multicol,lipsum}
\usepackage{amssymb} 
\usepackage{cuted}

\usepackage{amsmath,amssymb,amsfonts}

\usepackage{mleftright,xparse}

\NewDocumentCommand\xDeclarePairedDelimiter{mmm}
 {%
  \NewDocumentCommand#1{som}{%
   \IfNoValueTF{##2}
    {\IfBooleanTF{##1}{#2##3#3}{\mleft#2##3\mright#3}}
    {\mathopen{##2#2}##3\mathclose{##2#3}}%
  }%
 }
\xDeclarePairedDelimiter{\set}{\lbrace}{\rbrace}

\ifCLASSOPTIONcompsoc
  \usepackage[nocompress]{cite}
\else
  \usepackage{cite}
\fi
\ifCLASSINFOpdf
\else
\fi
\begin{document}
%
\title{Semi-supervised Learning with the EM Algorithm: A Comparative Study between Unstructured and Structured Prediction}
%
%
%
%

\author{Wenchong He,
        Zhe Jiang,~\IEEEmembership{Member,~IEEE,}
\IEEEcompsocitemizethanks{\IEEEcompsocthanksitem Corresponding author: Zhe Jiang, zjiang@cs.ua.edu\protect
\IEEEcompsocthanksitem W. He and Z. Jiang were with the Department
of Computer Science, University of Alabama, Tuscaloosa,
AL, 35487.}
\thanks{Manuscript received April 19, 2005; revised August 26, 2015.}}

\IEEEtitleabstractindextext{%
\begin{abstract}
Semi-supervised learning aims to learn prediction models from both labeled and unlabeled samples. There has been extensive research in this area. Among existing work, generative mixture models with Expectation-Maximization (EM) is a popular method due to clear statistical properties. However, existing literature on EM-based semi-supervised learning largely focuses on unstructured prediction, assuming that samples are independent and identically distributed. Studies on EM-based semi-supervised approach in structured prediction is limited. This paper aims to fill the gap through a  comparative study between unstructured and structured methods in EM-based semi-supervised learning. Specifically, we compare their theoretical properties and find that both methods can be considered as a generalization of self-training with soft class assignment of unlabeled samples, but the structured method additionally considers structural constraint in soft class assignment. We conducted a case study on real-world flood mapping datasets to compare the two methods. Results show that structured EM is more robust to class confusion caused by noise and obstacles in features in the context of the flood mapping application.
\end{abstract}

}

\maketitle

\IEEEdisplaynontitleabstractindextext

%
\IEEEpeerreviewmaketitle

\IEEEraisesectionheading{\section{Introduction}\label{sec:intro}}
\IEEEPARstart{S}emi-supervised learning aims to learn prediction models based on both labeled and unlabeled samples. It is important when training data contains limited labeled samples but abundant unlabeled samples. In real world spatial prediction problems, input data often contains abundant explanatory features but very limited ground truth. For example, in earth image classification for land cover mapping, a large number of image pixels with spectral features are collected from remote sensing platforms, but only a limited number of pixels are labeled with ground truth land cover classes~\cite{jiang2017spatial,shekhar2015spatiotemporal}. The reason is that collecting ground truth is both expensive and time consuming by field crew or well-trained image interpreters. 

The topic of semi-supervised learning has been extensively studied in the literature. According to a survey~\cite{zhu2005semi}, techniques can be categorized into generative  models with EM method, graph-based methods, label propagation, self-training and co-training, etc. In the EM-based method~\cite{ghahramani1994learning}, unknown labels are considered as hidden variables, and both labeled and unlabeled samples are used to estimate parameters of a generative model of the joint distribution. 
Graph-based methods assume that samples connected by heavy edges tend to have the same label and thus aim to obtain a smooth label assignment over the graph~\cite{camps2007semi}. In label propagation, the main idea is to propagate the labels of nodes throughout the network and form class communities~\cite{zhuѓ2002learning}. Semi-supervised learning in structured output space focuses on learning dependency or some internal relationships between classes~\cite{brefeld2006semi}.  In self-training, a classifier is first trained with a small amount of labeled samples and then is used to classify unlabeled samples~\cite{dopido2013semisupervised}. The most confident predictions on unlabeled samples together with their predicted class labels are added to the training set. For co-training, it assumes that features can be split into two sets and each set is sufficient to train a good classifier separately~\cite{hong2015spatial}. Each classifier then classifies the unlabeled data and teaches the other classifier with the  most confident predictions on unlabeled samples~\cite{zhu2005semi}. 

Among all the methods, semi-supervised learning  with EM provides clear statistical properties. There are two types of EM algorithms: unstructured EM and structured EM. Unstructured EM assumes that data samples follow an identical and independent distribution and feature variables are independent~\cite{ghahramani1994learning}. It considers unknown class labels as hidden variables in generative mixture models(e.g., Gaussian mixture models). It then uses EM algorithm to learn model parameters and infer hidden classes at the same time. Structured EM assumes that either input samples or feature variables within samples follow structural dependency. Existing works on structured  EM can be further categorized into two types: Bayesian structural EM and Hidden Markov Model (HMM). Bayesian structural EM uses a directed acyclic graph (DAG) to represent the conditional dependency between feature variables within a sample. It uses EM to learn the DAG structure from samples whose feature variables are partially unobserved (treated as hidden variables) ~\cite{friedman2013bayesian}. HMM uses a chain~\cite{damian2018algorithm} or tree~\cite{Xie2018} to represent the conditional dependency between samples. It uses EM to learn the model parameters and infer hidden classes at the same time. Structured EM is of particular interest because data in many real-world applications (e.g., earth science, biology, and material science) often show structural constraints. However, there is limited study that compares structured and unstructured methods in EM-based semi-supervised learning. It remains unclear about their strengths and weaknesses.










  
  
  


























To fill the gap, this paper provides a comparative study between structured methods and unstructured methods in EM-based semi-supervised learning for spatial classification problems~\cite{jiang2018survey}. For unstructured methods, we use Gaussian mixture models as an example. For structured methods, we use a recent model called \emph{geographical hidden Markov tree (HMT)}~\cite{Xie2018} as a representative example.   We compare the theoretical properties and conducted detailed case studies on real-world datasets. In summary, this paper makes the following contributions:
\begin{itemize}
    \item We compared the theoretical properties of a representative  unstructured methods (Gaussian Mixture Model)  and structured methods (HMT) in EM-based semi-supervised learning. 
    
    \item Through theoretical analysis, we found out that both EM-based methods can be considered as a generalization of self-training with soft class assignment of unlabeled samples. The difference is that structured methods additionally consider structure constraints in soft class assignment.
    \item We also empirically compared the performance of a representative unstructured EM (Gaussian mixture model) and structured EM (hidden Markov tree) in a case study on real-world flood mapping datasets. Results showed that in this particular application, unstructured EM without spatial feature could be impacted by feature noise and obstacles. Adding spatial feature into unstructured EM could alleviate but not fully resolve the issue. In contrast, structured EM could resolve the issue due to explicit structural constraint on this particular type of applications
\end{itemize}


\section{Problem Formulation}
Suppose we have data set $\mathcal{D}$ of $N$ data samples, which is composed of an unlabeled subset $ \mathcal{D}_u$ and a labeled subset $\mathcal{D}_l$($\mathcal{D} = \mathcal{D}_l\cup\mathcal{D}_u$). $\mathcal{D}_u $ contains $N_u$ data samples of inputs without class labels, $\mathcal{D}_u = \{ (\mathbf{x}_n) \}_{n=1}^{N_u}$. The labeled subset $\mathcal{D}_l $ contains $N_l$ data samples of input-output pairs, $\mathcal{D}_l = \{ (\mathbf{x}_n, y_n) \}_{n=N_u + 1}^{N}$ where $\mathbf{x}_n\in \mathbb{R}^{m\times 1}$ is a vector of $m$ explanatory features (including both non-spatial features and spatial contextual feature) with each element corresponding to one feature layer, and $y_n\in \{0,1\}$ is a binary class label.

Explanatory features of labeled samples and unlabeled samples  are noted as
$\mathbf{X}_u=[\mathbf{x}_{1},...,\mathbf{x}_{N_u}]^T$ and $\mathbf{X}_l=[\mathbf{x}_{N_u + 1},...,\mathbf{x}_{N}]^T$  respectively. The class layers are noted as $\mathbf{Y}_u=[y_{1},...,y_{N_u}]^T$ and $\mathbf{Y}_l=[y_{N_u + 1},...,y_{N}]^T. \mathbf{Y}_u$ is unknown and to be predicted. 
For example, in flood mapping from earth imagery, $\mathbf{x}_n$ can be the spectral band values (and elevation) of each pixel, and $y_n$ is the class label (flood, dry). We also assume $N_u \gg N_l$. Labeled subset $\mathcal{D}_l$ and an unlabeled subset $\mathcal{D}_u$ are training and test samples respectively. Note that both $\mathbf{X}_l$ and $\mathbf{X}_u$ are used in training (semi-supervised learning). 

\section{EM Algorithms for Semi-supervised Learning}
This section compares theoretical properties of EM-based semi-supervised learning algorithms in unstructured and structured prediction respectively.

\subsection{Example of Unstructured EM: Gaussian Mixture Models}
EM-based semi-supervised learning algorithm in unstructured prediction does not consider structural dependency of sample classes and assumes that the sample classes are independent and identically distributed (i.i.d.). 

The joint distribution of all samples is as below:
\begin{equation}\label{eq:empostexpll}
    \begin{split}
    P(\mathbf{X},\mathbf{Y})&=P({\mathbf{X}_u},\mathbf{Y}_u,\mathbf{X}_l,\mathbf{Y}_l) \\
    &=\prod_{n=1}^{N_u} P({\mathbf{x}_n},y_n)
    \prod_{n=N_u+1}^{N} P({\mathbf{x}_n},y_n)\\
    \end{split}
\end{equation}

We assume that the sample features follow an i.i.d. Gaussian distribution in each class. The prior class probability $P(y_n = c)$ follows a Bernoulli distribution below.
\begin{equation}\label{eq:empriordis}
    P(y_n = c) = \pi_c, c = \{0, 1\} .
\end{equation}
Feature distribution $P(\mathbf{x}_n|y_n)$  is as below,
\begin{equation}\label{eq:unfeatureclassprob}
        P(\mathbf{x}_{n}|y_n = c)\sim \mathcal{N}(\boldsymbol{\mu}_{c},\mathbf{\Sigma}_{c})
\end{equation}

We denote the entire set of parameters as $\mathbf{\Theta}=\{ \pi_c, \boldsymbol{\mu}_c, \mathbf{\Sigma}_c|c=0,1 \}$. The labeled data subset $\mathcal{D}_l $ is used to initialize parameters $\mathbf{\Theta}$. Then EM algorithm creates a function for the posterior expectation of log-likelihood evaluated using current estimate for the parameters(E-Step) and next updates parameters maximizing the expected log-likelihood for the next iteration (M-Step). The unknown class $\mathbf{Y}_u$ is hidden variable. The posterior expectation of log likelihood of all samples  is as below:
\begin{equation}\label{eq:empostexpll}\small
    \begin{split}
&\mathbb{E}_{\mathbf{Y}_u|\mathbf{X},\mathbf{Y}_l,\mathbf{\Theta_0}}\log P({\mathbf{X}_u},\mathbf{Y}_u,\mathbf{X}_l,\mathbf{Y}_l|\mathbf{\Theta})\\
    &=\mathbb{E}_{\mathbf{Y}_u|\mathbf{X},\mathbf{Y}_l,\mathbf{\Theta_0}}\log( \prod_{n=1}^{N_u} P({\mathbf{x}_n},y_n|\mathbf{\Theta}) \prod_{n=N_u+1}^{N} P({\mathbf{x}_n},y_n|\mathbf{\Theta}) )\\
    &=\sum\limits_{\mathbf{Y}_u} P(\mathbf{Y}_u|\mathbf{X},\mathbf{Y}_l,\mathbf{\Theta_0})(\sum_{n=1}^{N_u}\log{P({\mathbf{x}_n},y_n|\mathbf{\Theta})} + \sum\limits_{n=N_u + 1}^{N}\log{P({\mathbf{x}_n},y_n|\mathbf{\Theta})})\\
    & =\sum_{n=1}^{N_u}\sum\limits_{y_n} P(y_n|\mathbf{x}_n,\mathbf{\Theta_0})\log{P({\mathbf{x}_n},y_n|\mathbf{\Theta})} \\
    &\quad\quad\quad+ \sum_{n=N_u + 1}^{N}\sum\limits_{y_n} P(y_n|\mathbf{x}_n,\mathbf{Y}_l,\mathbf{\Theta_0})\log{P({\mathbf{x}_n},y_n|\mathbf{\Theta})}\\
    & =\sum_{n=1}^{N_u}\sum\limits_{y_n} P(y_n|\mathbf{x}_n,\mathbf{\Theta_0})\log{P({\mathbf{x}_n},y_n|\mathbf{\Theta})} \\
    &\quad\quad\quad+\sum_{n=N_u + 1}^{N}\sum\limits_{c}\mathbf{I}(y_n=c)\log{P({\mathbf{x}_n},y_n=c|\mathbf{\Theta})}\\
    \end{split}
\end{equation}

The  posterior class distribution for unlabeled samples ($1\leq n\leq N_u$)
$P({y}_n|{\mathbf{x}_n},\mathbf{\Theta_0})$ is

\begin{equation}\label{eq:emposteriordis}\small
    \begin{split}
        P({y}_n = c|{\mathbf{x}_n},\mathbf{\Theta_0}) = \frac{P({\mathbf{x}_n}|y_n = c,\mathbf{\Theta_0})P(y_n = c)}{\sum\limits_{c'=0}^1P({\mathbf{x}_n}|y_n=c',\mathbf{\Theta_0})P(y_n=c')}, c = {0, 1}
    \end{split}
\end{equation}
 
Taking the above into the posterior expectation of log likelihood, we can easily get the following formulas  to update parameters that maximize the posterior expectation (M-Step). Note that $\mathbf{I}(y_n=c) = 1$ if $y_n=c$, and 0 otherwise.

\begin{equation}\label{eq:priorupdate}\small
    \pi_c=\frac{ \sum\limits_{n=1}^{N_u}P(y_n = c|{\mathbf{x}_n},\mathbf{\Theta_0} ) + \sum\limits_{n=N_u+1}^{N}\mathbf{I}(y_n = c)}{ \sum\limits_{c'=0}^1\sum\limits_{n=1}^{N_u}P(y_n = c'|{\mathbf{x}_n},\mathbf{\Theta_0}) + \sum\limits_{c'=0}^1\sum\limits_{n=N_u+1}^{N}\mathbf{I}(y_n = c') }
\end{equation}

\begin{equation}\label{eq:muupdate}\small
        \mu_c=\frac{ \sum\limits_{n=1}^{N_u}P(y_n = c|{\mathbf{x}_n},\mathbf{\Theta_0} ){\mathbf{x}_n} + \sum\limits_{n=N_u+1}^{N}\mathbf{I}(y_n=c){\mathbf{x}_n} }{ \sum\limits_{n=1}^{N_u}P(y_n = c|{\mathbf{x}_n},\mathbf{\Theta_0}) + \sum\limits_{n=N_u+1}^{N}\mathbf{I}(y_n=c) }
\end{equation}

\begin{equation}\label{eq:sigmaupdate}\scriptscriptstyle
        \Sigma_c=\frac{ \sum\limits_{n=1}^{N_u}P(y_n = c|{\mathbf{x}_n},\mathbf{\Theta_0} )({\mathbf{x}_n} - \mu_c)({\mathbf{x}_n} - \mu_c)^T + \sum\limits_{n=N_u+1}^{N}\mathbf{I}(y_n=c)({\mathbf{x}_n} - \mu_c)({\mathbf{x}_n} - \mu_c)^T }{ \sum\limits_{n=1}^{N_u}P(y_n = c|{\mathbf{x}_n},\mathbf{\Theta_0}) + \sum\limits_{n=N_u+1}^{N}\mathbf{I}(y_n=c) }
\end{equation}

\emph{Class inference}: After learning model parameters, we can infer hidden class variables by maximizing the log joint probability of unlabeled data.

\begin{equation}\label{eq:unjointdis}
    \begin{split}
        \log P(\mathbf{X}_u,\mathbf{Y}_u) &= \log \prod_{n=1}^{N_u}P({\mathbf{x}_n}|y_n)P(y_n)\\ 
        &= \sum_{n=1}^{N_u}\log P({\mathbf{x}_n}|y_n)P(y_n)
    \end{split}
\end{equation}

To maximize the total log probability,  we can maximize each term in Equation~\ref{eq:unjointdis}. For each sample n, we simply choose the class c that gives higher probability of $P(\mathbf{x}_n|y_n=c)P(y_n=c)$.

\subsection{Example of Structured EM: Hidden Markov Tree}
EM-based semi-supervised learning algorithm in  structured prediction  assumes a dependency structure between sample classes. We use a spatial classification model   called  \emph{geographical hidden Markov tree (HMT)}\cite{Xie2018}. It is a probabilistic graphical model that generalizes the common hidden Markov model (HMM) from a one-dimensional sequence to a partial order tree. 

The joint distribution of unlabeled samples' features and classes can be formulated as Equation~\ref{eq:empostall}, where ${\mathcal{P}_n}$ is the set of parent samples of the $n$th sample in the dependency tree (${\mathcal{P}_n}=\emptyset$ for a leaf node), and $y_{k\in\mathcal{P}_n}\equiv\{y_k|k\in\mathcal{P}_n\}$ is the set of parent node classes of node $n$.
\begin{equation}\label{eq:empostall}
    \begin{split}
    P(\mathbf{X}_u,\mathbf{Y}_u)
    =\prod_{n=1}^{N_u} P({\mathbf{x}_n}|y_n) \prod_{n=1}^{N_u}P(y_n|y_{k\in\mathcal{P}_n})
    \end{split}
\end{equation}

Similar to EM for unstructured prediction, the model assumes that features in each class follows an i.i.d. Gaussian distribution.
\begin{equation}\label{eq:featureclassprob}
        P(\mathbf{x}_{n}|y_n = c)\sim \mathcal{N}(\boldsymbol{\mu}_{c},\mathbf{\Sigma}_{c})
\end{equation}

Class transitional probability follows the partial order flow dependency constraint\cite{Xie2018}, as shown in Table~\ref{tab:transitionprob}.
\begin{table}[h]
\caption{Class transition probability and prior probability}
\label{tab:transitionprob}
\begin{tabular}{|c|c|c|}\hline
$P(y_n|y_{\mathcal{P}_n})$ & $y_{\mathcal{P}_n}=0$ & $y_{\mathcal{P}_n}=1$\\ \hline
$y_n=0$ & $1$ & $1-\rho$\\ \hline
$y_n=1$ & $0$ & $\rho$ \\ \hline
\end{tabular}\hspace{2mm}
\begin{tabular}{|c|c|c|}\hline
 & $P(y_n)$\\ \hline
$y_n=0$ & $\pi_0$\\ \hline
$y_n=1$ & $\pi_1$ \\ \hline
\end{tabular}
\end{table}

We denote the entire set of parameters as $\mathbf{\Theta}=\{\rho, \pi_c, \boldsymbol{\mu}_c, \mathbf{\Sigma}_c|c=0,1 \}$. The posterior expectation of log likelihood of unlabeled samples (E-step) is as below: 

\begin{equation}\label{eq:postexpll}\footnotesize
\begin{split}
&\mathbb{E}_{\mathbf{Y}_u|\mathbf{X}_u,\mathbf{\Theta_0}}\log P(\mathbf{X}_u,\mathbf{Y}_u|\mathbf{\Theta})\\
&=\mathbb{E}_{\mathbf{Y}_u|\mathbf{X}_u,\mathbf{\Theta_0}}\log\left\{ \prod_{n=1}^{N_u} P({\mathbf{x}_n}|y_n,\mathbf{\Theta}) \prod_{n=1}^{N_u}P(y_n|y_{k\in\mathcal{P}_n},\mathbf{\Theta})\right\}\\
&=\sum\limits_{\mathbf{Y}_u}{ P(\mathbf{Y}_u|\mathbf{X}_u,\mathbf{\Theta_0})}\\
&\quad\quad{\left\{\sum_{n=1}^{N_u}\log{P({\mathbf{x}_n}|y_n,\mathbf{\Theta})}+\sum_{n=1}^{{N_u}}\log{P(y_n|y_{k\in\mathcal{P}_n},\mathbf{\Theta})}\right\}}\\
& =\sum_{n=1}^{N_u}\sum_{y_n}P({y}_n|\mathbf{X},\mathbf{\Theta_0})\log{P({\mathbf{x}_n}|y_n,\mathbf{\Theta})}\\
&\quad\quad+\sum_{n=1}^{{N_u}}~\sum_{y_n,y_{k\in\mathcal{P}_n}}P(y_n,y_{k\in\mathcal{P}_n}|\mathbf{X},\mathbf{\Theta_0})\log{P(y_n|y_{k\in\mathcal{P}_n},\mathbf{\Theta})} \\
\end{split}
\end{equation}



After computation of marginal class posterior distribution through forward and backward message propagation, we can get the parameter updating formula by maximizing the posterior expectation of log likelihood as below (M-Step). 
\begin{equation}\label{eq:updaterho}\footnotesize
   \rho = \frac{\sum\limits_{n|\mathcal{P}_n\neq\emptyset}{\sum\limits_{y_n}\sum\limits_{y_{\mathcal{P}_n}}{y_{\mathcal{P}_n}(1 - y_n)P(y_n, y_{\mathcal{P}_n}|\mathbf{X}, \mathbf{\Theta_0})}}} 
 {\sum\limits_{n|\mathcal{P}_n\neq\emptyset}{\sum\limits_{y_n}\sum\limits_{y_{\mathcal{P}_n}}{y_{\mathcal{P}_n}P(y_n, y_{\mathcal{P}_n}|\mathbf{X}, \mathbf{\Theta_0})}}} 
\end{equation}

\begin{equation}\label{eq:updatepi}\footnotesize
  \pi_1 = \frac{\sum\limits_{n|\mathcal{P}_n=\emptyset}{\sum\limits_{y_n }{y_n P(y_n|\mathbf{X}, \mathbf{\Theta_0})}}} {\sum\limits_{n|\mathcal{P}_n=\emptyset}{\sum\limits_{y_n}{P(y_n|\mathbf{X}, \mathbf{\Theta_0})}}}  
\end{equation}

\begin{equation}\label{eq:updatemu}\footnotesize
    \mu_c = \frac{\sum\limits_{n} {\mathbf{x}_n} P(y_n = c|\mathbf{X},\mathbf{\Theta_0})} {\sum\limits_{n }  P(y_n = c|\mathbf{X},\mathbf{\Theta_0})},c = {0, 1}
\end{equation}

\begin{equation}\label{eq:updatesigma}\footnotesize
    \Sigma_c = \frac{\sum\limits_{n} ({\mathbf{x}_n} - \boldsymbol{\mu}_c) ({\mathbf{x}_n} - \boldsymbol{\mu}_c)^T P(y_n = c|\mathbf{X},\mathbf{\Theta_0})} {\sum\limits_{n} P(y_n = c|\mathbf{X},\mathbf{\Theta_0})}, c = {0, 1}
\end{equation}

\emph{Class inference}: After learning model parameters, we can infer hidden class variables by maximizing the overall joint probability. 
\begin{equation}\label{eq:logp}
\log P(\mathbf{X},\mathbf{Y})= \sum_{n=1}^{N_u} \log P({\mathbf{x}_n}|y_n) + \sum_{n=1}^{N_u}\log P(y_n|y_{k\in\mathcal{P}_n})
\end{equation}
A naive approach that enumerate all combinations of class assignment is infeasible due to the exponential cost. We use a dynamic programming based method called \emph{max-sum}~\cite{rabiner1989tutorial}.
\subsection{Comparison}

\begin{table}[ht]
\centering
\caption{Comparison between unstructured EM (i.i.d.) and structured EM}
\label{tab:paraComp1}
\begin{tabular}{p{1.4cm}p{3.1cm}p{3cm}}
\hline
 & Gaussian mixture model & Hidden Markov tree \\ \hline
{Parameter}&{$\pi_c,\mu_c,\Sigma_c$}&{$\rho,\pi_c,\mu_c,\Sigma_c$}\\ \hline
{Posterior probability}&{$p(y_n|x_n)$, from Bayes theorem}&{$p(y_n|\mathbf{X})$, from message propagation}. \\ \hline
\\

 Feature probability &{$p(x_n|y_n)\sim\mathcal{N}(\boldsymbol{\mu}_c,\mathbf{\Sigma}_c$)}&{ $p(x_n|y_n)\sim\mathcal{N}(\boldsymbol{\mu}_c,\mathbf{\Sigma}_c)$}\\ \hline
 Class \\ probability&{$p(y_n=c) =\pi_c$}&{$p(y_n=c) =\pi_c$ for leaf nodes, $p(y_n|y_{k\in\mathcal{P}_n})$ for non-leaf nodes}\\ \hline
\end{tabular}
\end{table}
Table \ref{tab:paraComp1} summarizes the comparison of theoretical properties between the two EM methods. From the expression of joint probabilities, both methods assume that features in each class follows an i.i.d. Gaussian distribution. The difference lies in the class prior probability. For unstructured EM methods, the class prior probability follows i.i.d. Bernoulli distribution $p(y_n) \sim \mathbf{B}(1,\pi)$ while for the structured EM method, sample classes follow a dependency structure , which can be expressed by class transition probability $P(y_n|y_{k\in\mathcal{P}_n})$~\cite{Xie2018}. Moreover, both methods have similar formulas for  parameter update  where sample mean and covariance are reweighted by each sample's class posterior probability.  The difference lies in the way they compute the class posterior probability. For unstructured model, it is from Bayes theorem; while for structured model it is from message propagation,  considering the class dependency structure~\cite{Xie2018,jiang2019geographical,jiang2019hidden,sainju2020hidden}.

When analysing the theoretical properties of the two EM algorithms, we found that EM algorithms can be considered as a generalization of self-training\cite{zhuѓ2002learning}.  In self-training, a classifier is first trained with a small amount of labeled data. The classifier is then used to classify the unlabeled data. The most confident unlabeled points together with their predicted labels are added to the training set in the next iteration~\cite{zhu2005semi}. In contrast, EM algorithm first uses labeled samples to initialize model parameters, and then estimates the class posterior probability of each unlabeled sample. In the next iteration, it uses the class posterior probability as a weight to re-estimate model parameter (Equations~\ref{eq:updaterho},~\ref{eq:updatepi},~\ref{eq:updatemu}, and \ref{eq:updatesigma}). In summary, both self-training and EM algorithm use the labeled data to iteratively learn model parameters. The difference is that self-training makes a hard class  assignment of unlabeled samples to retrain model in iterations while the EM algorithm uses the class posterior probability to make a soft class assignment for each unlabeled sample.

\begin{figure}
    \vspace{-1mm}
    \centering
    \includegraphics[height=0.9in]{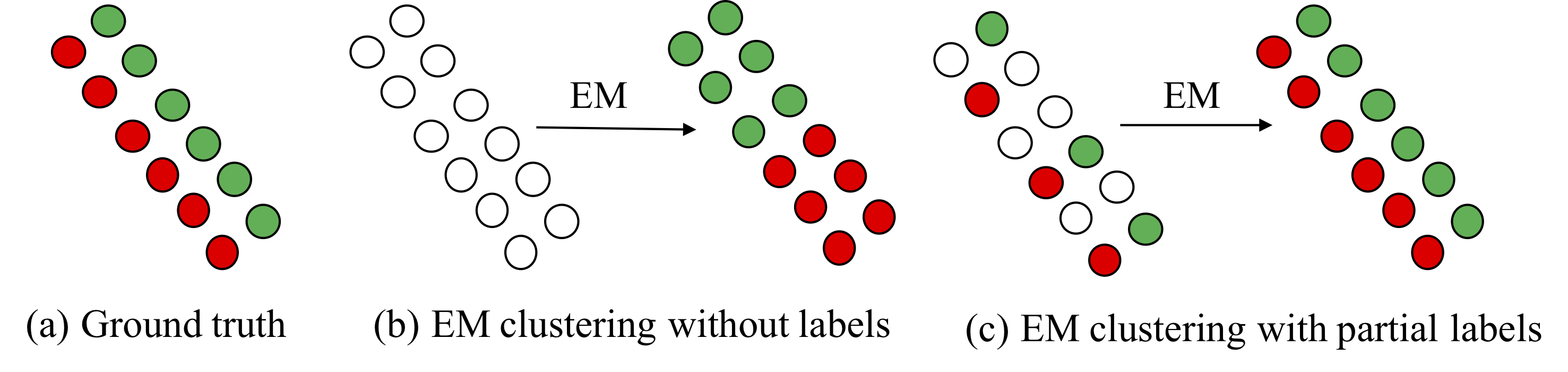}
    \caption{Illustration of importance of labeled samples in Gaussian mixture models (red and green are two classes)}
    \label{fig:emlabel}
   \end{figure}

It is important to note that the local optimal problem exists for both approaches. If class estimation of unlabeled data is misleading, it may further hurt learning in iterations. This potential problem can be alleviated by a good initial parameters estimation from representative training samples. The labeled samples are used to initialize some model parameters, i.e., the mean vectors $\mu_c$ and covariance matrices $\Sigma_c$ of features of samples in each class. These parameters could have been initialized randomly without labeled samples (unsupervised setting of Gaussian mixture model, also called EM clustering, as well as unsupervised setting of hidden Markov models).  However, randomly initialized parameters of mean and covariance matrix may not converge to good values after EM iterations, particularly when the feature clusters of samples in two classes are not well-separated from each other (as the case of our flood mapping application), as shown by the example in Figure~\ref{fig:emlabel}(b). In this situation, labeled samples in each class can be used to estimate its corresponding mean vector and covariance matrix more accurately, as shown by Figure~\ref{fig:emlabel}(c). 

    \begin{table}[ht]
    \centering
    \caption{Summary of unstructured EM and   structured EM in unsupervised learning
    and semi-supervised learning.}
    \label{tab:semicomp}
    \includegraphics[height=2.0in]{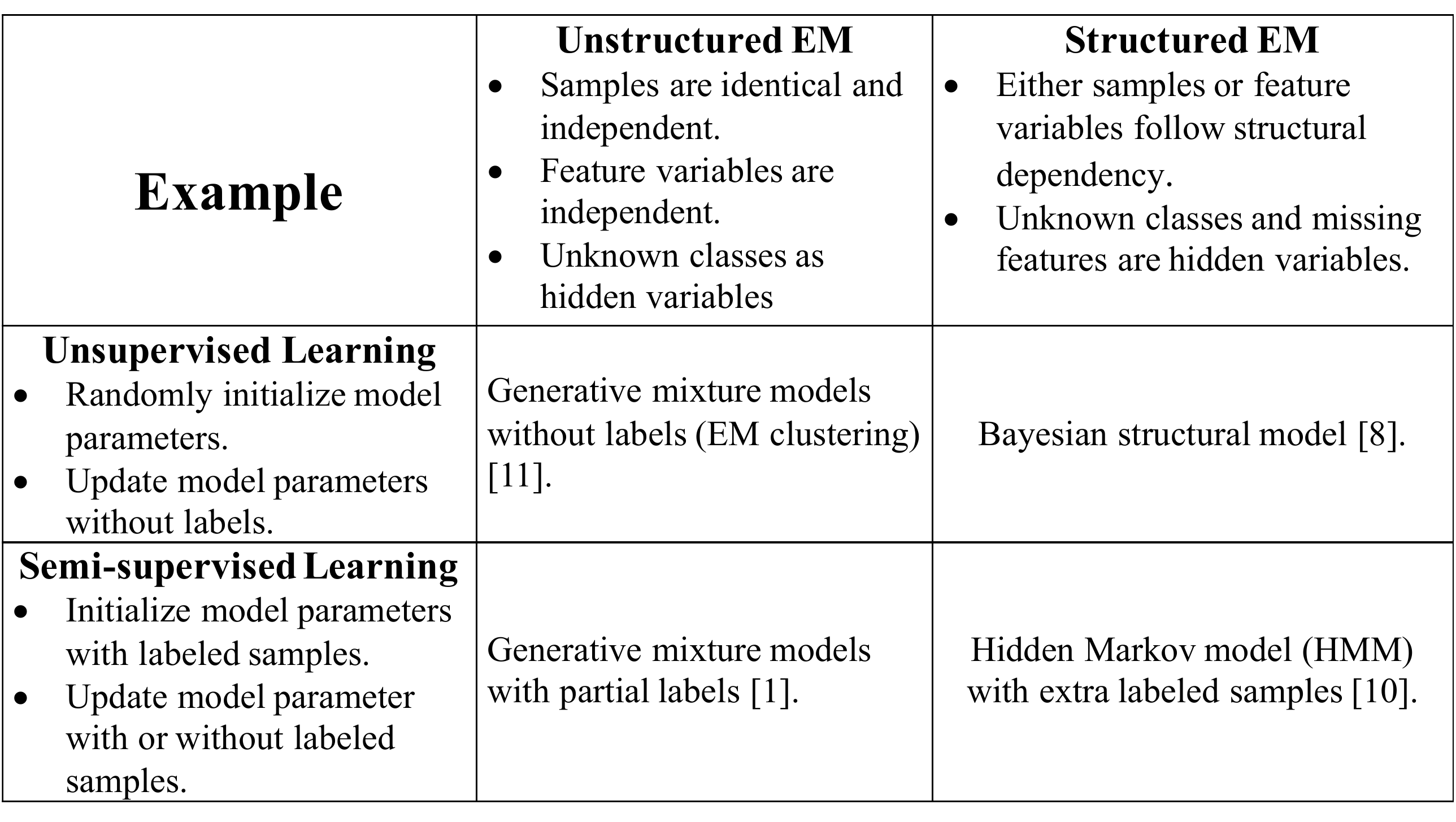}
   \end{table}
Table~\ref{tab:semicomp} compares unstructured EM versus structured EM in both unsupervised and semi-supervised settings in a broader perspective. Both unstructured and structured EM learn model parameters with the existence of missing variables (e.g., hidden class labels or missing feature observations). The difference is that structured EM incorporates structural dependency between samples or feature variables within a sample. In unsupervised setting, class labels are either fully unknown (e.g. , EM clustering) or not of relevance (e.g., feature dependency learning in Bayesian structured EM). Thus, initialization and update of model parameters do not rely on class labels. In semi-supervised setting, class labels are partially available. These labels can be used in parameter initialization and potentially in parameter update as well. Specifically, in Gaussian mixture models with partial labels, both labeled and unlabeled samples are used in parameter initialization as well as class inference. In hidden Markov models, if extra labeled samples are available, these labels can be used to provide a reasonable initialization of some model parameters (e.g., mean and covariance matrix in each class). After this, model parameters are iteratively updated based on the features and dependency structure on test samples. The process belongs to transdutive learning since a model is learned for a specific structure across test samples. It is worth to note that HMMs can also be unsupervised with randomly initialized parameters without class labels. But the converged parameters in this case may be ineffective in discriminating two classes on test samples (see~\cite{Xie2018}).

\section{Evaluation}\label{sec:eval}
In this section, we compared the performance of unstructured EM prediction with structured EM prediction through case studies on real world datasets. Our goal is to get insights on how well different EM methods can handle class confusion due to noise and obstacles in features. 
Moreover, we also compare EM algorithm with other baseline methods. We chose Gradient Boost Machine and Random forest models as baseline because these models have well-tested source codes and have also shown superior performance over other models in the literature. The candidate classification methods are listed below.  
\begin{itemize}
    \item {\bf Unstructured EM (Gaussian mixture model) w/o elevation feature}: We implemented our codes in Matlab.

    \item {\bf Unstructured EM (Gaussian mixture model) w/ elevation feature }: We implemented our codes in Matlab.
\item {\bf Structured EM (HMT)}: We implemented the HMT source code in C++.
\item {\bf Gradient Boost Model (GBM)}: We used the GBM in R gbm packages on raw features together with elevation feature. 
\item {\bf Random forest (RF) }: We used the random forest in R randomForest packages on raw features together with elevation feature.
\end{itemize}

\emph{Hyperparameter:}  
For unstructured EM, the hyper-parameter includes the parameter convergence threshold, and the cutoff threshold to decide the positive and negative classes. 
For structured EM, the hyper-parameter includes the parameter convergence threshold, and initial parameter values of $\rho$ (class transitional probability) and $\pi$ (class prior probability). The parameter convergence threshold was set as 1e-5, and the cutoff threshold was set as 0.5. We set $\rho=0.99$ and $\pi=0.5$ based on earlier sensitivity study (see ~\cite{Xie2018}). For random forest, the hyper-parameters include the number of trees $N_1$, the number of variables randomly sampled as candidates at each split $N_2$ and the minimum size of terminal nodes $N_3$. For $N_1$, we tried values of 300, 350, 400, 450, 500. For $N_2$, we tried values of 1, 2, 3. For $N_3$, we tried values of 5, 10, 20, 40. When choosing the optimal value for one parameter, we kept other parameters constant. The optimal values are: $N_1 = 350$, $N_2= 2$ and $N_3 = 10$.  For gradient boosted model, the hyper-parameters include the number of trees $T_1$, the maximum depth of each tree $T_2$, and the shrinkage parameter $T_3$ that is used to reduce the impact of each additional fitted tree.  For $T_1$, we tried values of 1000, 1500, 2000, 2500, 3000. For $T_2$, we tried values of 1, 2, 3, 4. For $T_3$, we tried values of 0.1, 0.01, 0.001. The optimal hyper-parameters are: $T_1= 2000$, $T_2= 1$, and $T_3 = 0.01$. 

\begin{figure}
       \centering
       \subfloat[High-resolution satellite imagery in NC.]{%
            \includegraphics[width=2.7in]{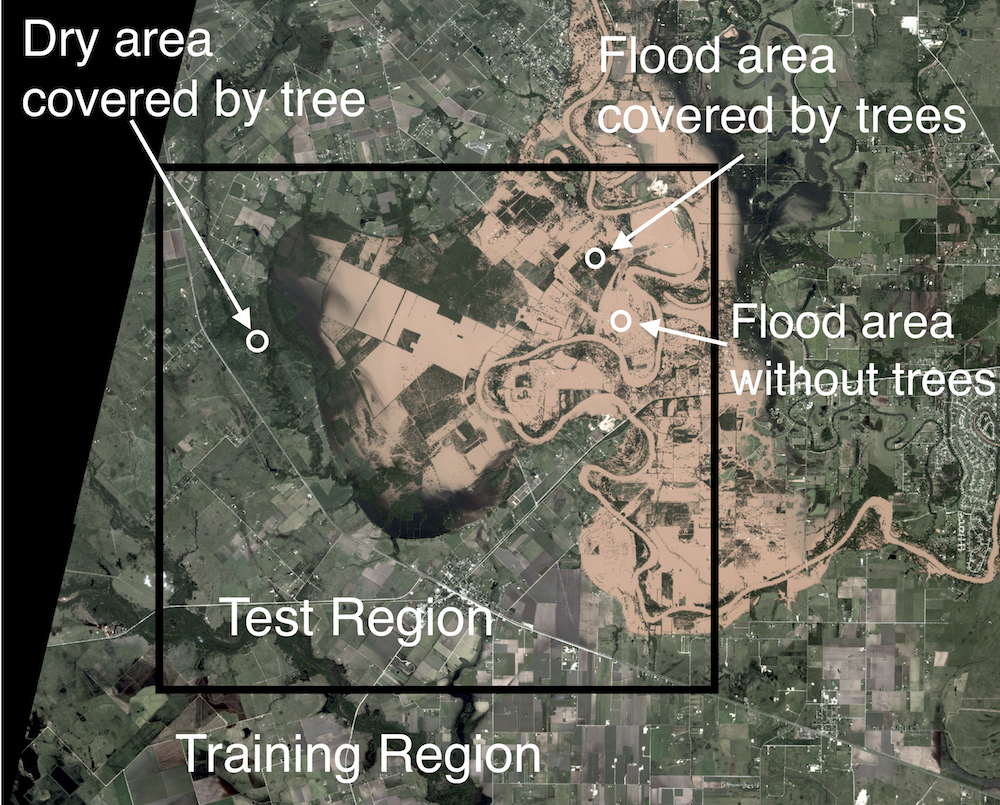}
       }\\
        \subfloat[Digital elevation]{%
       \includegraphics[width=2.7in]{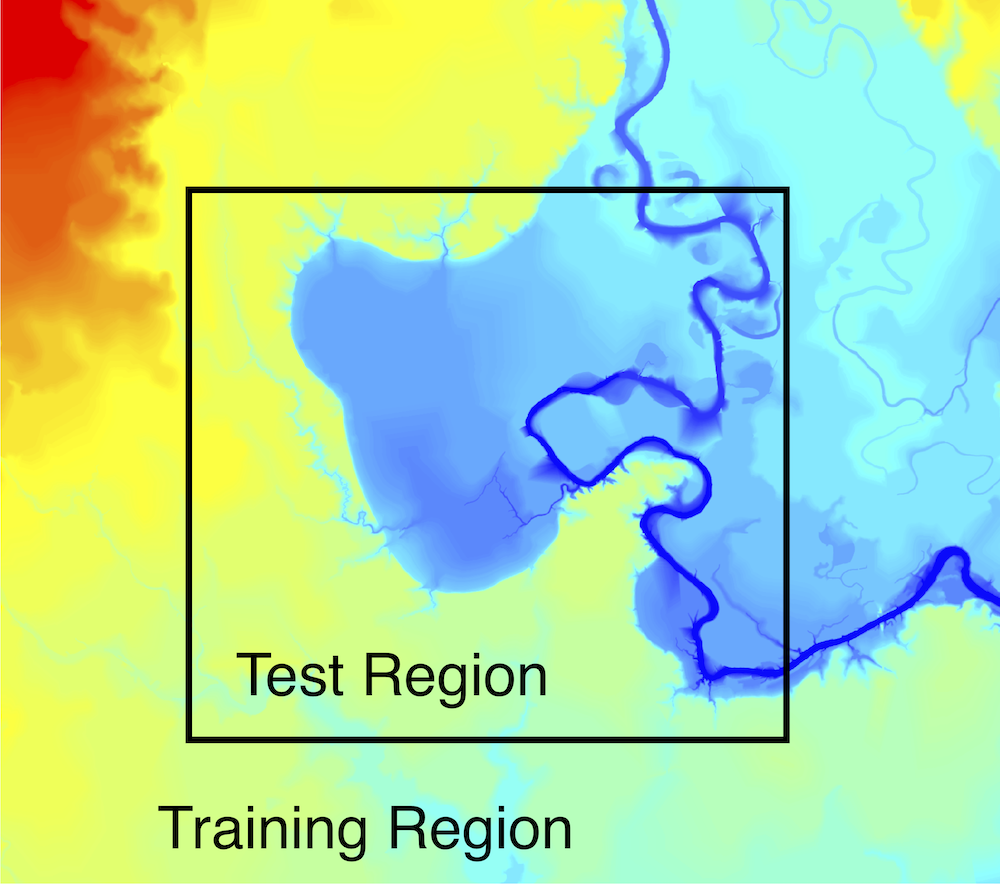}
       }
       \caption{RGB feature and spatial elevation feature}
       \label{fig:InputFeature}
   \end{figure}

\emph{Dataset description:}
We used two flood mapping datasets from Hurricane Harvey floods in Texas 2017 and Hurricane Mathew in North Carolina 2016 respectively. 
Non-spatial explanatory features include red, green, blue bands in aerial imagery from PlanetLab Inc. and NOAA National Geodetic Survey~\cite{ngs} respectively. The spatial contextual feature was digital elevation map from Texas natural resource management department and the University of North Carolina Libraries~\cite{ncsudem} respectively. All data were resampled into 2 meter by 2 meter resolution. Figure~\ref{fig:InputFeature} shows the entire input features in the Harvey dataset, including non-spatial features (RGB bands) in Figure~\ref{fig:InputFeature}(a) and spatial contextual feature (elevation) in Figure~\ref{fig:InputFeature}(b). From the images, we can see class confusion due to noise and obstacles in non-spatial features (there are pixels with tree colors in both flood and dry areas). Due to space limit, we put the results on the second dataset in Appendix.

    \begin{figure}
    \vspace{-2mm}
    \centering
    \includegraphics[height=2.7in]{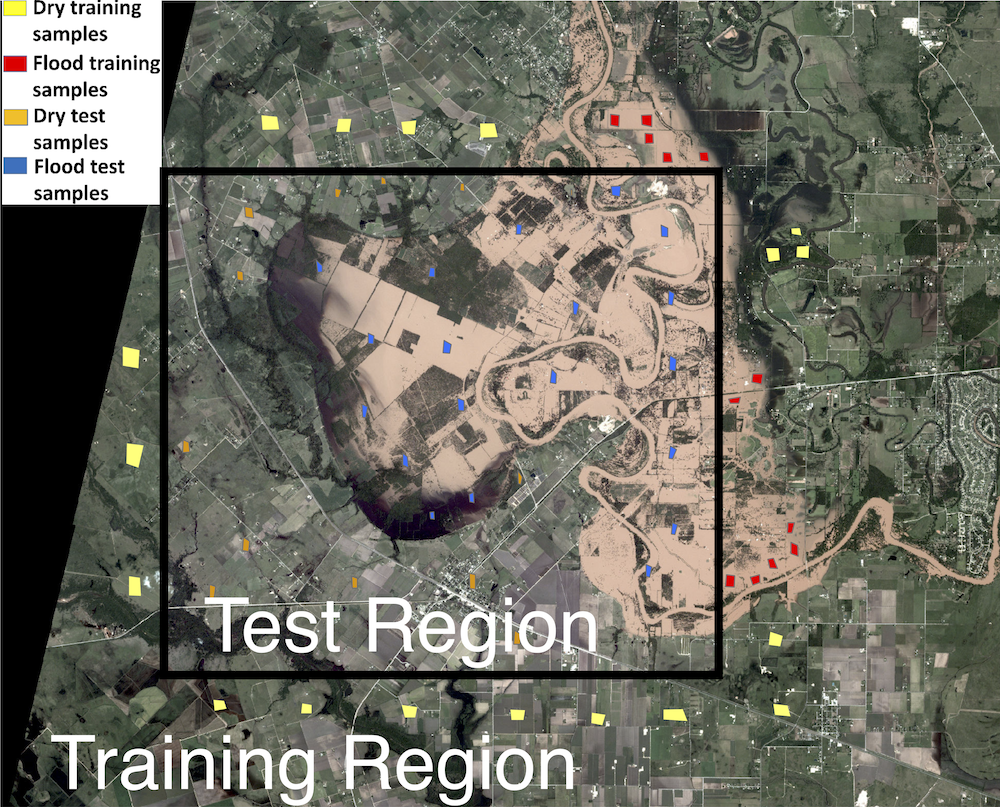}
    \caption{Training and test polygons}
    \label{fig:datasplit}
   \end{figure}
   
\emph{Training and test dataset split:}
We used simple validation. The separation of training and test sets is shown in Figure~\ref{fig:datasplit}. We had a test region (highlighted by the black rectangle) with labeled polygons in both classes. The training region with training polygons in both classes was outside the test region. In the experiment, we randomly selected 10000 pixels from training polygons (5000 in flood and 5000 in dry) and 103374 pixels from test polygons (43972 in dry and 59402 in flood)

\begin{figure}
\centering
\subfloat[Iteration of parameter $\boldsymbol{\mu}$]{%
      \includegraphics[height=1.5in,width=1.7in]{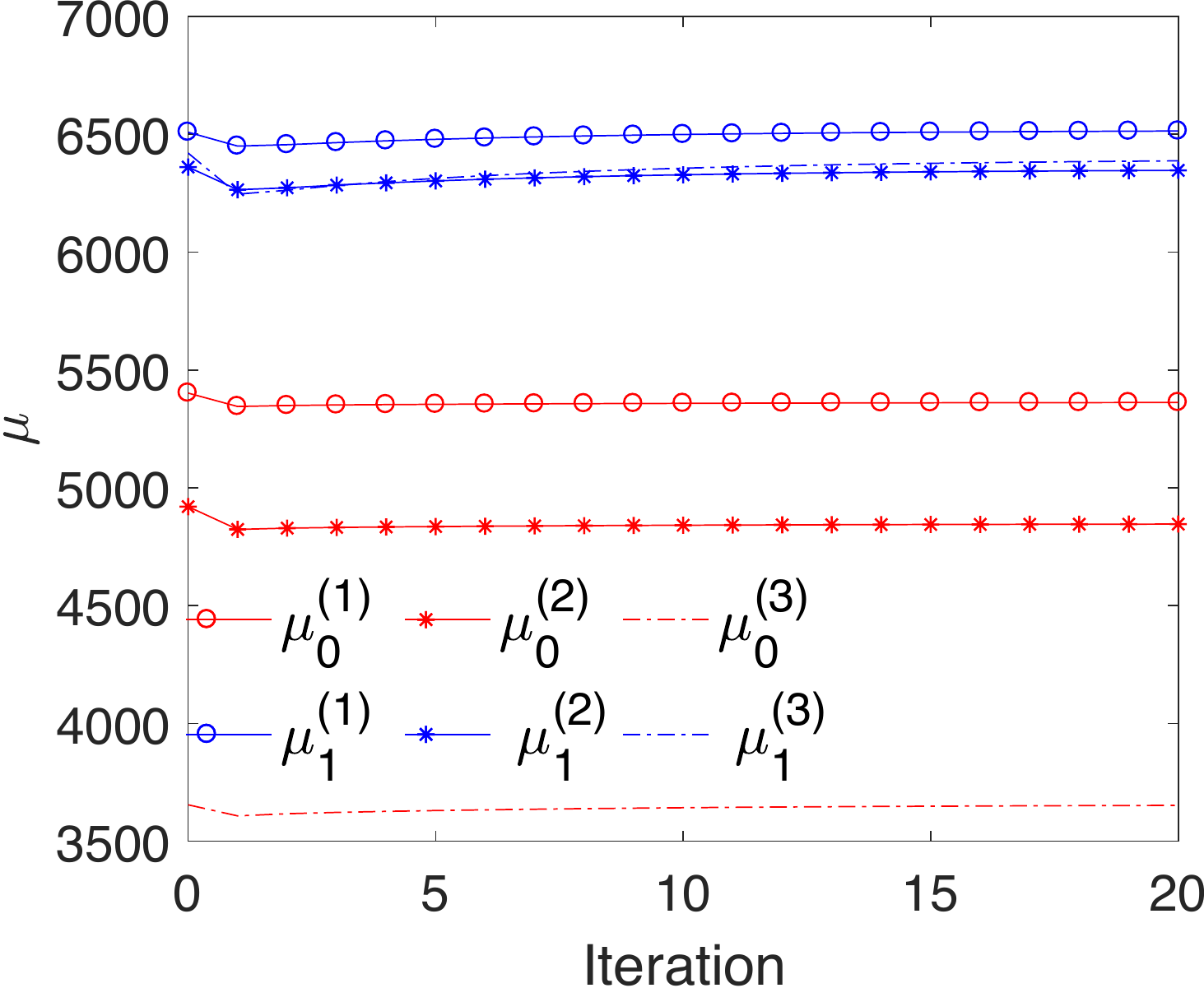}
}
\subfloat[Iteration of parameter $\boldsymbol{\Sigma}$]{%
      \includegraphics[height=1.6in,width=1.6in]{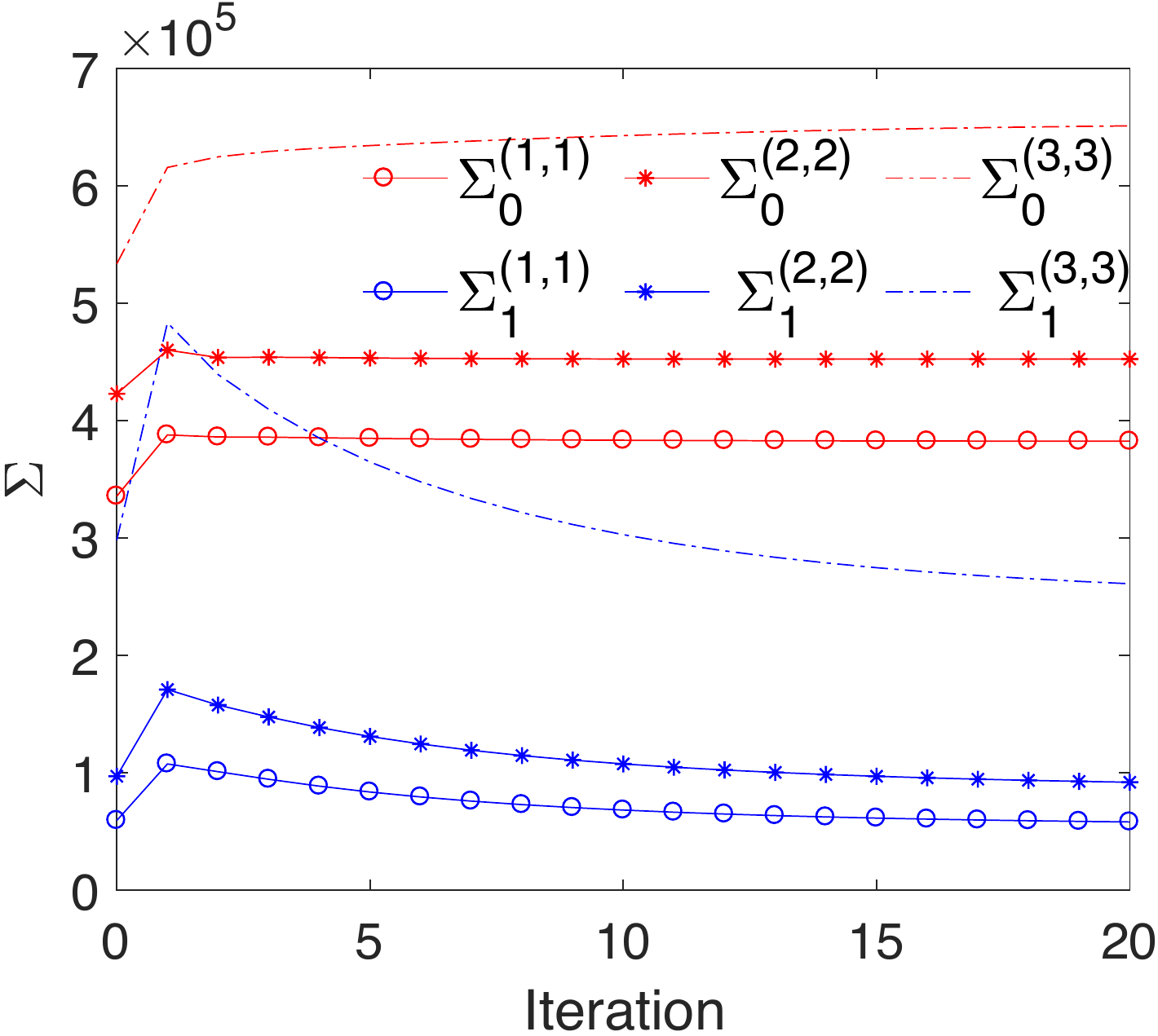}
}
\caption{Parameter iterations and convergence for unstructured EM without elevation feature}
\label{fig:EMNOElevmuSigmaSec}
\end{figure}

\begin{figure}
\vspace{-5mm}
\centering
\subfloat[Iteration of parameter $\boldsymbol{\mu}$]{%
      \includegraphics[height=1.5in,width=1.7in]{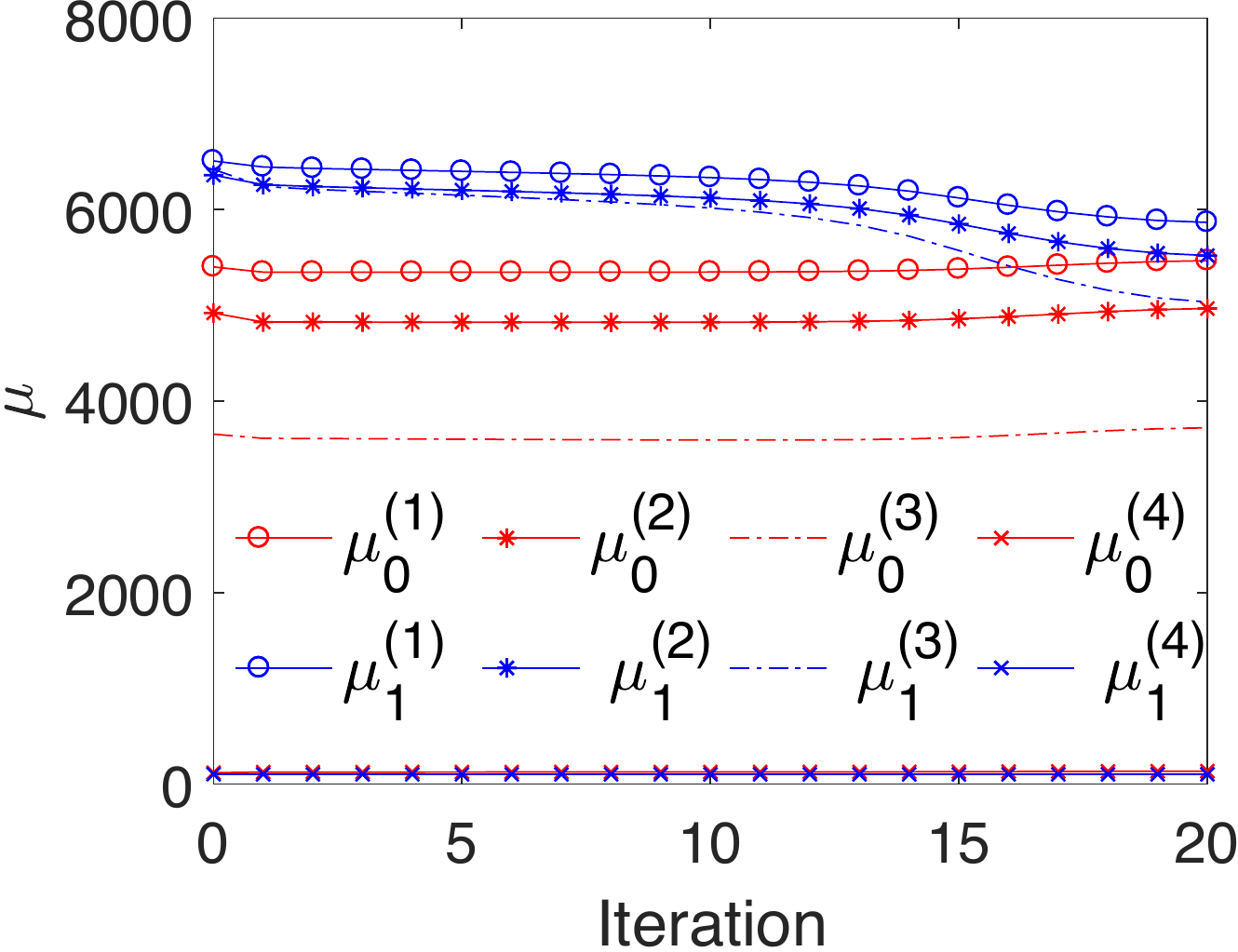}
}
\subfloat[Iteration of parameter $\boldsymbol{\Sigma}$]{%
      \includegraphics[height=1.6in,width=1.6in]{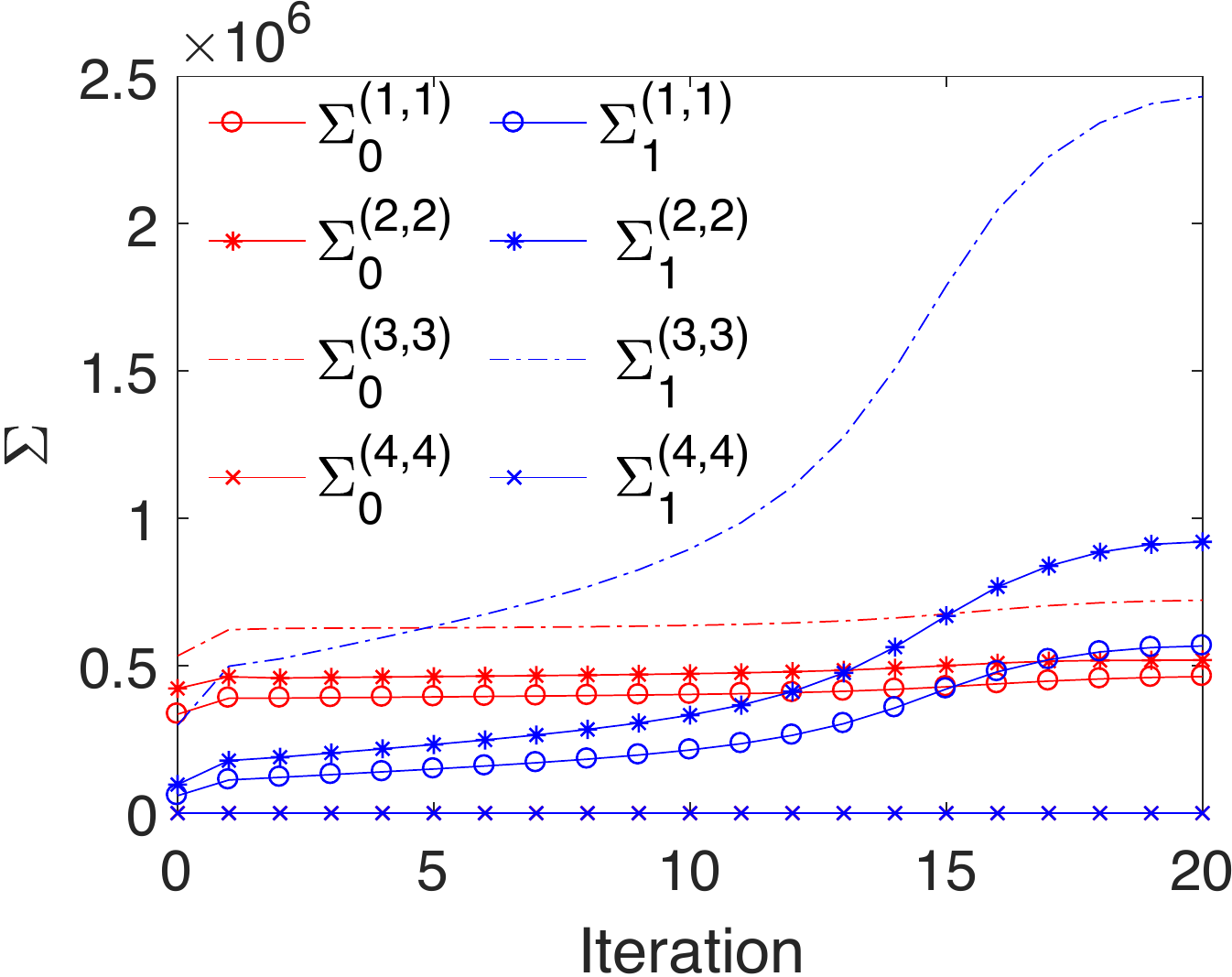}
}
\caption{Parameter iterations and convergence for unstructured EM with elevation feature}
\label{fig:EMmuSigmaSec}
\end{figure}

\begin{figure}
\vspace{-5mm}
\centering
\subfloat[Iteration of parameter $\boldsymbol{\mu}$]{%
      \includegraphics[height=1.5in,width=1.7in]{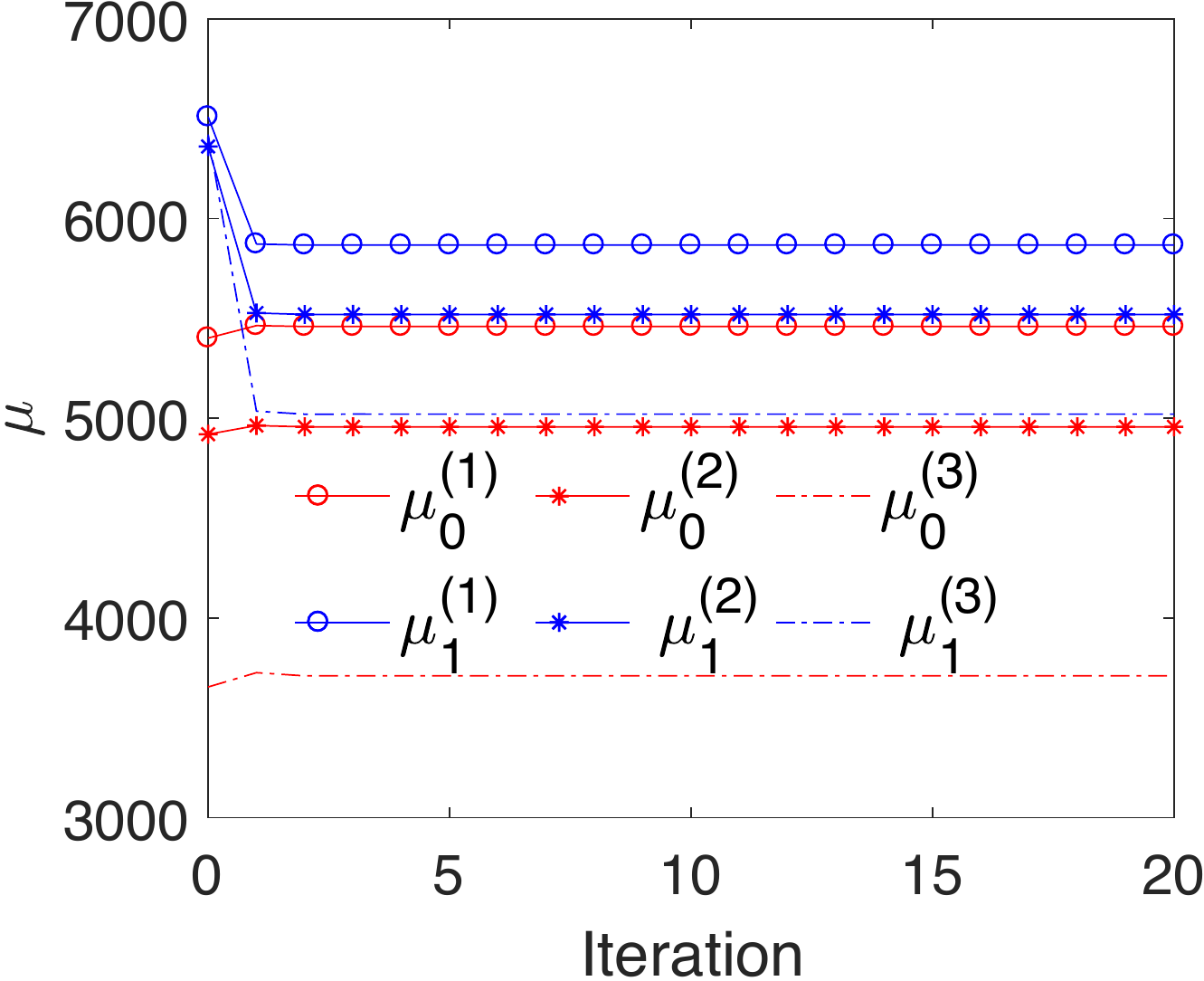}
}
\subfloat[Iteration of parameter $\boldsymbol{\Sigma}$]{%
      \includegraphics[height=1.6in,width=1.6in]{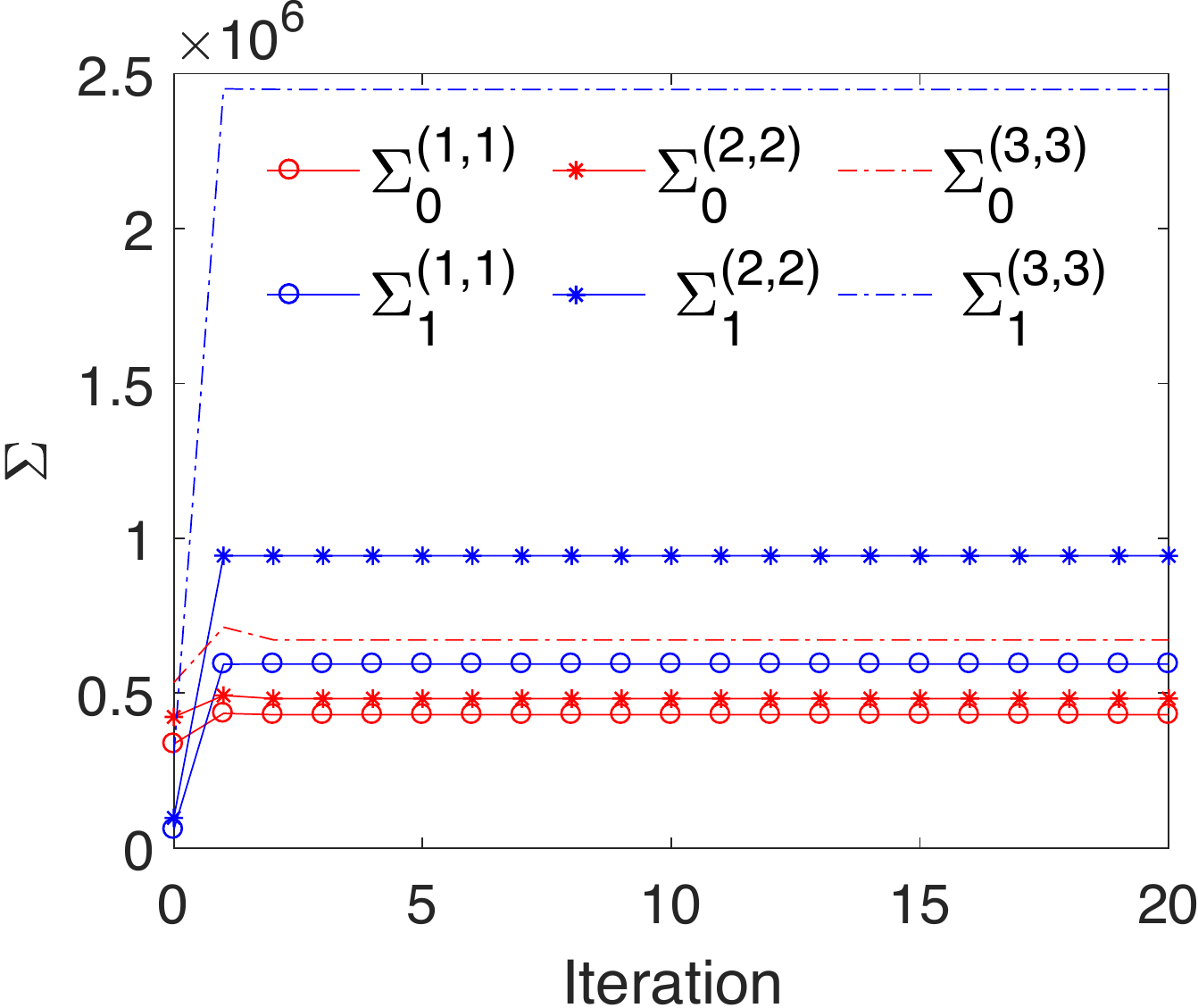}
}
\caption{Parameter iterations and convergence for structured EM algorithm}
\label{fig:HMTmuSigmaSec}
\end{figure}
 
\begin{figure*}
\centering
\subfloat[Satellite imagery]{%
      \includegraphics[height=0.95in, width=1.0in]{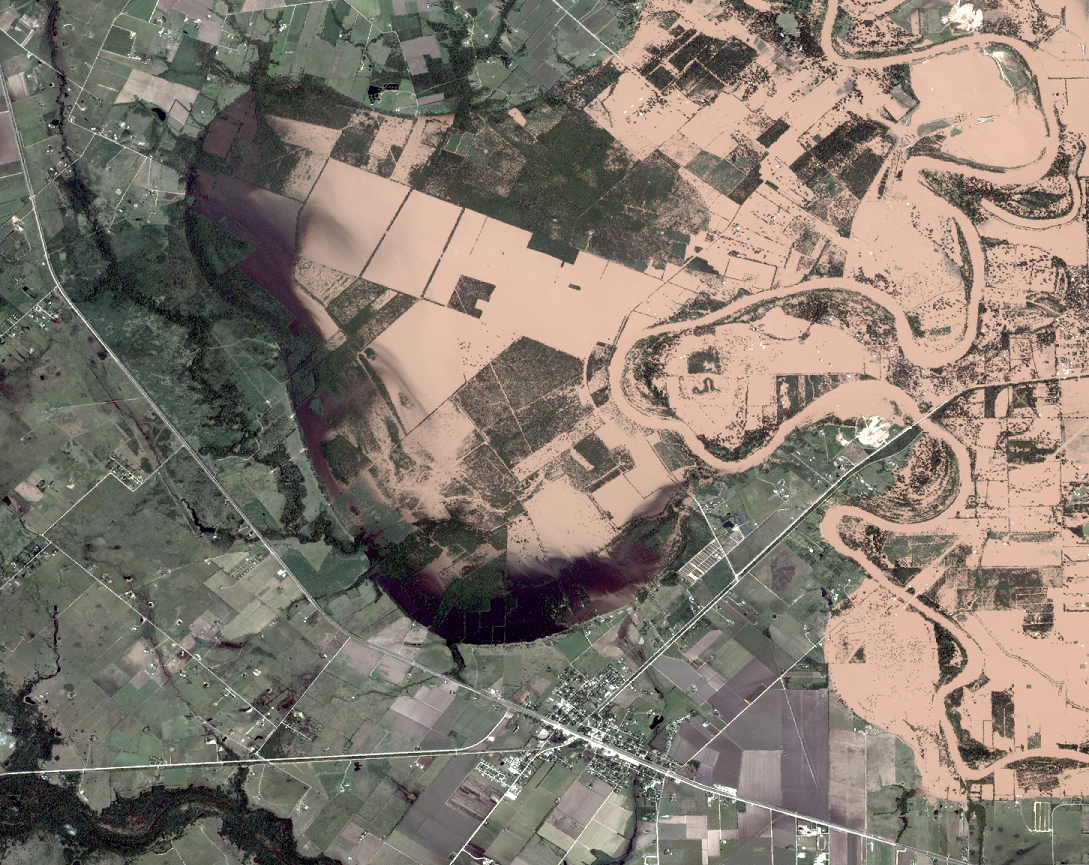}
}
\subfloat[Iteration 0]{%
      \includegraphics[height=1.0in]{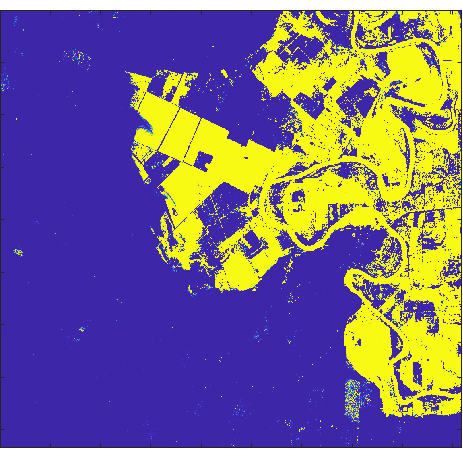}
}
\subfloat[Iteration 5]{%
      \includegraphics[height=1.0in]{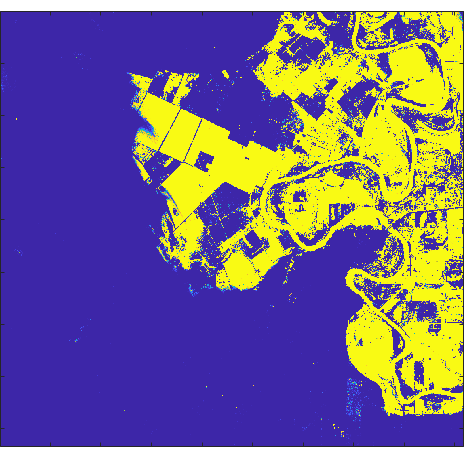}
}
\subfloat[Iteration 10]{%
      \includegraphics[height=1.0in]{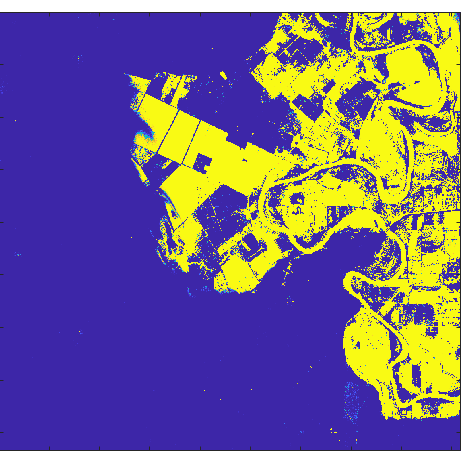}
}
\subfloat[Iteration 15]{%
      \includegraphics[height=1.0in]{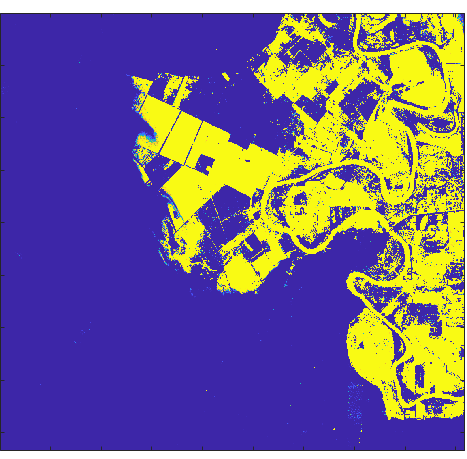}
}
\subfloat[Iteration 20]{%
      \includegraphics[height=1.0in]{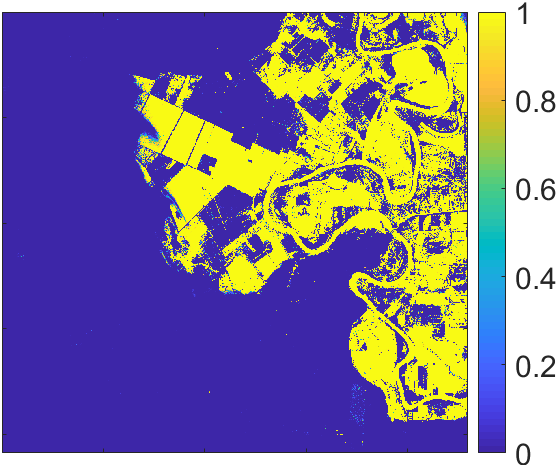}
}

\vspace{-2mm}
\caption{Posterior Probability of Unstructured EM without elevation}
\label{fig:EMNOelevPosterProbSec}
\vspace{-5mm}
\end{figure*}

\begin{figure*}
\centering
\subfloat[Satellite imagery]{%
      \includegraphics[height=0.95in, width=1.0in]{figures/true.png}
}
\subfloat[Iteration 0]{%
      \includegraphics[height=1.0in]{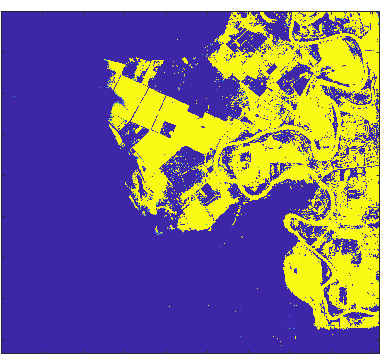}
}
\subfloat[Iteration 5]{%
      \includegraphics[height=1.0in]{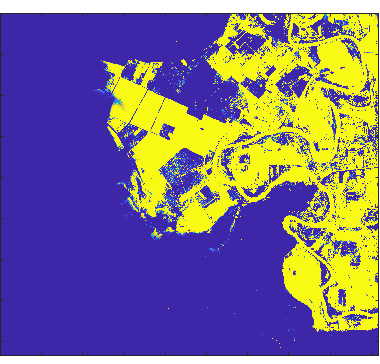}
}
\subfloat[Iteration 10]{%
      \includegraphics[height=1.0in]{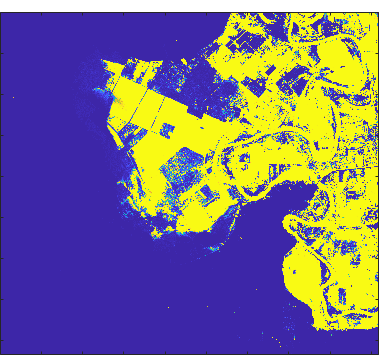}
}
\subfloat[Iteration 15]{%
      \includegraphics[height=1.0in]{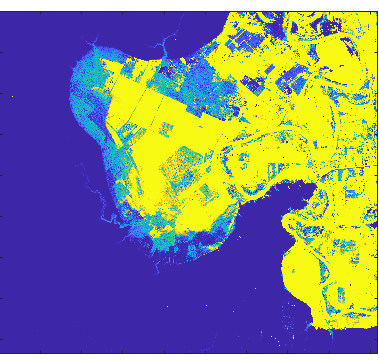}
}
\subfloat[Iteration 20]{%
      \includegraphics[height=1.0in]{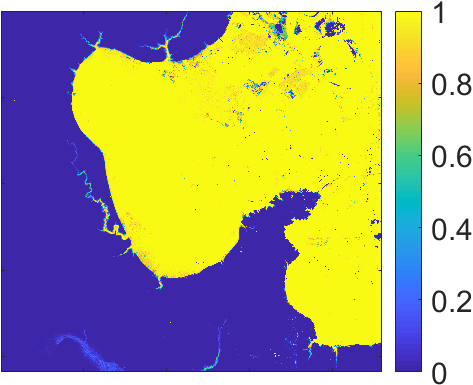}
}

\vspace{-2mm}
\caption{Posterior Probability of Unstructured EM with elevation}
\label{fig:EMwithelevPosterProbSec}
\vspace{-5mm}
\end{figure*}

\begin{figure*}
\centering
\subfloat[Satellite imagery]{%
      \includegraphics[height=0.95in, width=1.0in]{figures/true.png}
}
\subfloat[Iteration 0]{%
      \includegraphics[height=1.0in]{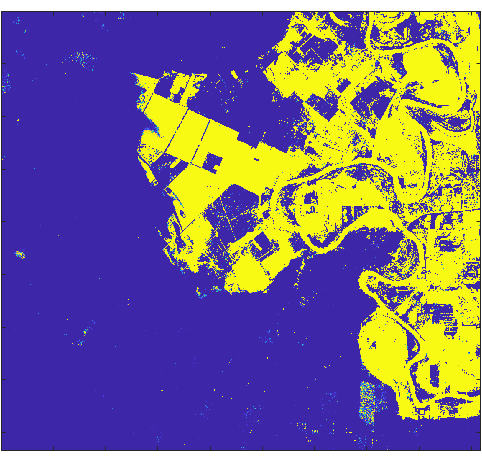}
}
\subfloat[Iteration 5]{%
      \includegraphics[height=1.0in]{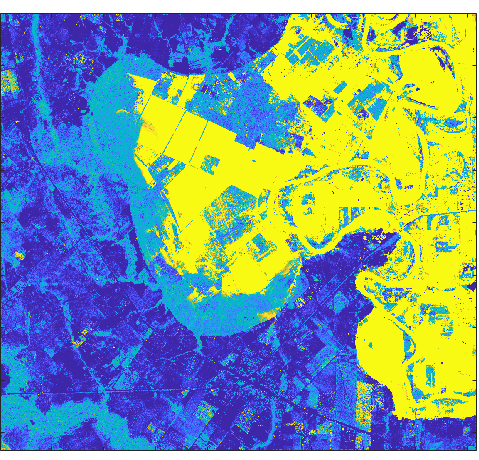}
}
\subfloat[Iteration 10]{%
      \includegraphics[height=1.0in]{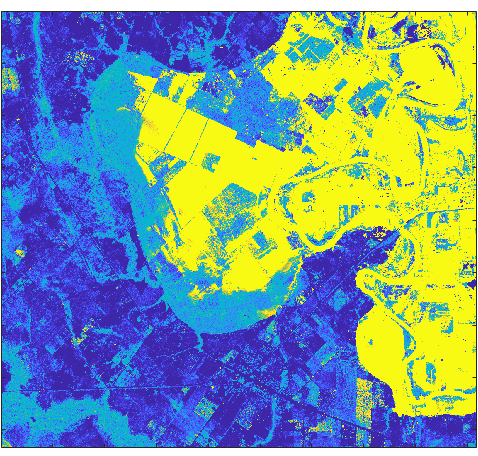}
}
\subfloat[Iteration 15]{%
      \includegraphics[height=1.0in]{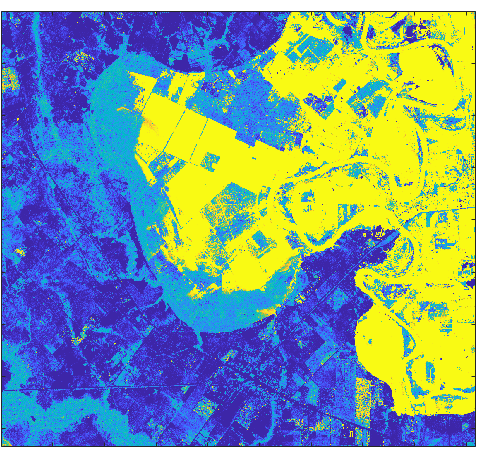}
}
\subfloat[Iteration 20]{%
      \includegraphics[height=1.0in]{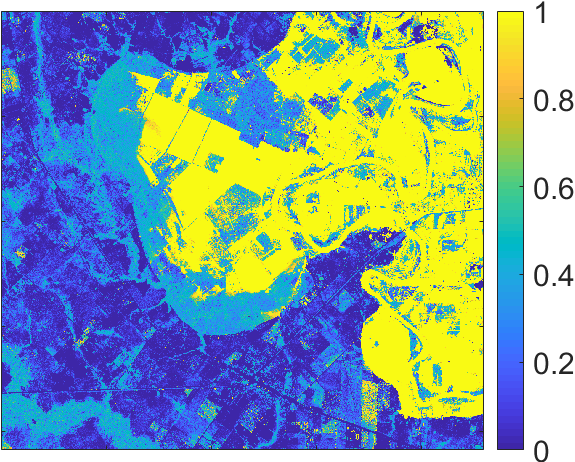}
}
\vspace{-2mm}
\caption{Posterior Probability of structured EM}
\label{fig:HMTPosterProbSec}
\vspace{-5mm}
\end{figure*}

\subsection{Parameter Convergence}
{\bf Parameter iteration and convergence:} Our convergence threshold was set to $0.001\%$. Figure \ref{fig:EMNOElevmuSigmaSec} shows the iterations of  $\boldsymbol{\mu}_c,\boldsymbol{\Sigma}_c$ ($c=0$ for the dry class, $c=1$ for the flood class) in unstructured EM without spatial contextual feature (elevation). For $\boldsymbol{\Sigma}_c$, we only plotted the diagonal  elements (i.e., the variance of each feature) and omitted covariance values due to space limit. From the results, we can see that the converged mean values of the two classes are well separated with $\boldsymbol{\mu}_0$ converged to $3500$ to $5500$ and $\boldsymbol{\mu}_1$ converged to $6300$ to $6500$. This is consistent with Figure~\ref{fig:InputFeature}(a) since flood areas have lighter color than dry areas. Note that the range of values in red, blue, and green bands are bigger than 256 due to a different imagery data type.
We can also observe that the variance of the flood class $\boldsymbol{\Sigma}_1$ converged to a lower range of values (around $1\times 10^5\sim 3\times 10^5$). The reason is that unstructured EM without spatial contextual feature will re-group samples with class confusion in their feature values (e.g., tree pixels with the same color in both flood and dry areas) into the same class based on class posterior (Figure~\ref{fig:EMNOelevPosterProbSec}), which in turn will influence the update of parameters in each class.

For unstructured EM with spatial contextual feature (elevation), the parameter iteration and convergence are shown in Figure~\ref{fig:EMmuSigmaSec}. Note that there is one more dimension for elevation in the plots. We can see that the converged mean values of two classes are less separately compared with Figure~\ref{fig:EMNOElevmuSigmaSec}. $\boldsymbol{\mu}_0$ converges to a range of $3800$ to $5500$ and $\boldsymbol{\mu}_1$ converges to a range of $5000$ to $6000$. Another dramatical change is on the variance of flood class $\boldsymbol{\Sigma_1}$, which increases to a much larger range of values ($0.5\times 10^6$ to $2.5\times 10^6$). This can be explained by the marginal class posterior probabilities in Figure~\ref{fig:EMwithelevPosterProbSec}, where samples with high posterior probability in the flood class (yellow color) are growing and include more tree pixels in the flood areas. Because of this, the variance of the flood class grows bigger and the mean of the flood class drops (tree pixels in water are darker than exposed flood water).

For structured EM, the converged parameter values are moderate in the middle of the above two cases, but more similar to unstructured EM with spatial contextual feature (elevation). The results can be explained by the posterior class probability in Figure~\ref{fig:HMTPosterProbSec}, where samples with high posterior probability in the flood class (yellowish pixels) are moderately in between the previous two maps. 
The main difference is that $\boldsymbol{\mu}_c$ and $\boldsymbol{\Sigma}_c$ converge in fewer  (only 2) iterations compared with unstructured EM with elevation.

\subsection{Posterior Class Probability}
Figures~\ref{fig:EMNOelevPosterProbSec},~\ref{fig:EMwithelevPosterProbSec}, and~\ref{fig:HMTPosterProbSec} show the iterations of posterior probabilities of samples in the flood class for unstructured EM without and with elevation features as well as structured EM respectively. Sample class posterior probability is important in understanding how the EM algorithm works because it shows how much a sample contribute to the parameter update for the next iteration in each class (e.g., Equations~\ref{eq:priorupdate},~\ref{eq:muupdate},~\ref{eq:sigmaupdate}, and Equations~\ref{eq:updaterho},~\ref{eq:updatepi},~\ref{eq:updatemu},~\ref{eq:updatesigma}). For unstructured EM, posterior class probability is estimated based on the Bayes Theorem with an i.i.d. assumption. Thus, samples with class confusion in feature values will be estimated towards the same class. This explains why pixels correspond to trees in the flood have low posterior probability in the flood class in Figure~\ref{fig:EMNOelevPosterProbSec} (same as the trees in dry areas). In contrast, unstructured EM with the elevation feature could separate confused tree pixels in flood and dry areas since their elevation values differ. This is shown in Figure~\ref{fig:EMwithelevPosterProbSec}(f). Finally, for structured EM, posterior class probability is estimated based on both local class likelihood from non-spatial features (RGB colors) and class dependency structure. This explains why the class posterior in Figure~\ref{fig:HMTPosterProbSec} is moderate compared with Figure~\ref{fig:EMNOelevPosterProbSec} and Figure~\ref{fig:EMwithelevPosterProbSec}. It is worth noting that posterior class probability for individual pixels are not the same as final class prediction in structured EM (final class prediction could be smoother due to jointly predicting all classes with dependency).

\subsection{Prediction Result}

\begin{figure}
\centering
\subfloat[Unstructured EM without elevation]{%
\includegraphics[height=1.2in]{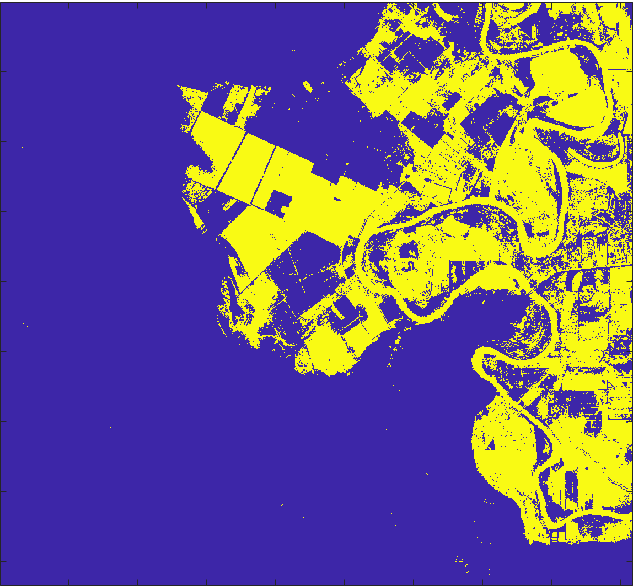} 
}
\subfloat[Unstructured EM with elevation]{%
    \includegraphics[height=1.2in]{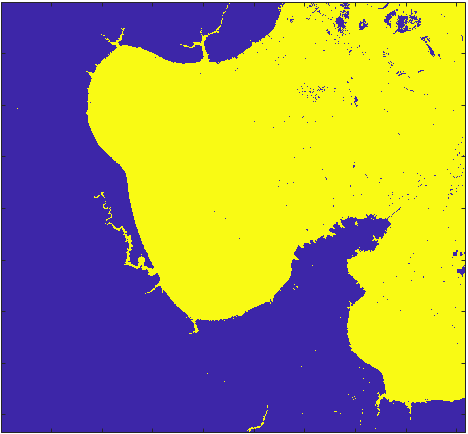} 
    }\\
\subfloat[Structured EM]{%
    \includegraphics[height=1.2in]{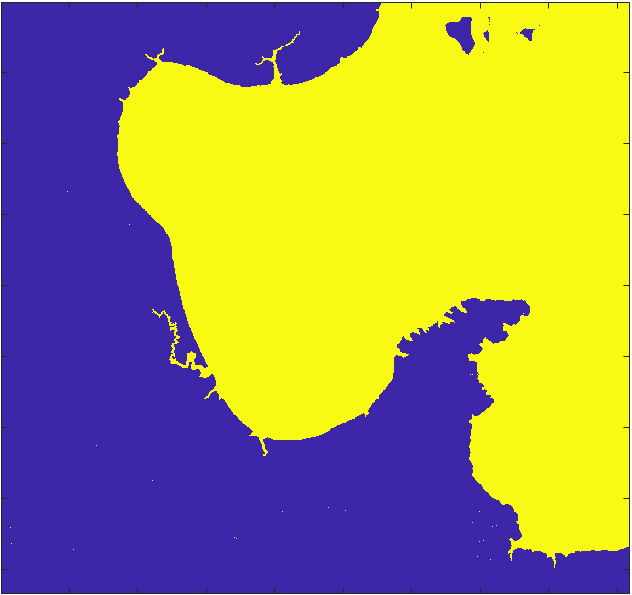} 
    }
\subfloat[Satellite imagery]{%
      \includegraphics[height=1.2in,width=1.25in]{figures/true.png}
}
\caption{Comparison of class prediction}
\label{fig:predresultSec}
\end{figure}

\begin{table}[h]\footnotesize
\centering
\caption{Comparison on  Harvey, Texas flood data}
\label{tab:comp1}
\begin{tabular}{cccccc}
\hline
Classifiers & Class & Prec. &Recall & F & Avg. F\\ \hline
\multirow{2}{*}{GBM }&Dry&{0.88}&{0.98}&{0.93}&\multirow{2}{*}{0.87}\\ 
 &Flood&{0.96}&{0.70}&{0.81}&\\ \hline
\multirow{2}{*}{RF}&Dry&{0.70}&{0.99}&{0.82}&\multirow{2}{*}{0.82}\\ 
 &Flood&{0.99}&{0.69}&{0.81}&\\ \hline
\multirow{2}{*}{Unstructured EM w/o elev.}&Dry&{0.68}&{0.99}&{0.81}&\multirow{2}{*}{0.75}\\ 
 &Flood&{0.99}&{0.53}&{0.70}&\\ \hline
\multirow{2}{*}{Unstructured EM w/ elev.}&Dry&{0.99}&{0.99}&{0.99}&\multirow{2}{*}{0.99}\\ 
 &Flood&{0.99}&{0.99}&{0.99}&\\ \hline 
\multirow{2}{*}{EM Structured}&Dry&{0.99}&{0.99}&{0.99}&\multirow{2}{*}{0.99}\\ 
 &Flood&{0.99}&{0.99}&{0.99}&\\ \hline  
\end{tabular}
\end{table}
Table~\ref{tab:comp1} and Figure \ref{fig:predresultSec} show the final classification results of the three methods. We can see that unstructured EM without spatial contextual feature (elevation) performed poorly with class confusion. Unstructured EM with spatial contextual feature performs significantly better with less class confusion, but also produces some salt-and-pepper noise errors since spatial features are used with an i.i.d. assumption. In contrast, the structured EM method can both address   the class confusion issue and show a smooth class map due to explicitly considering spatial dependency structure.

\begin{figure}
    \vspace{-1mm}
    \centering
    \includegraphics[height=2in]{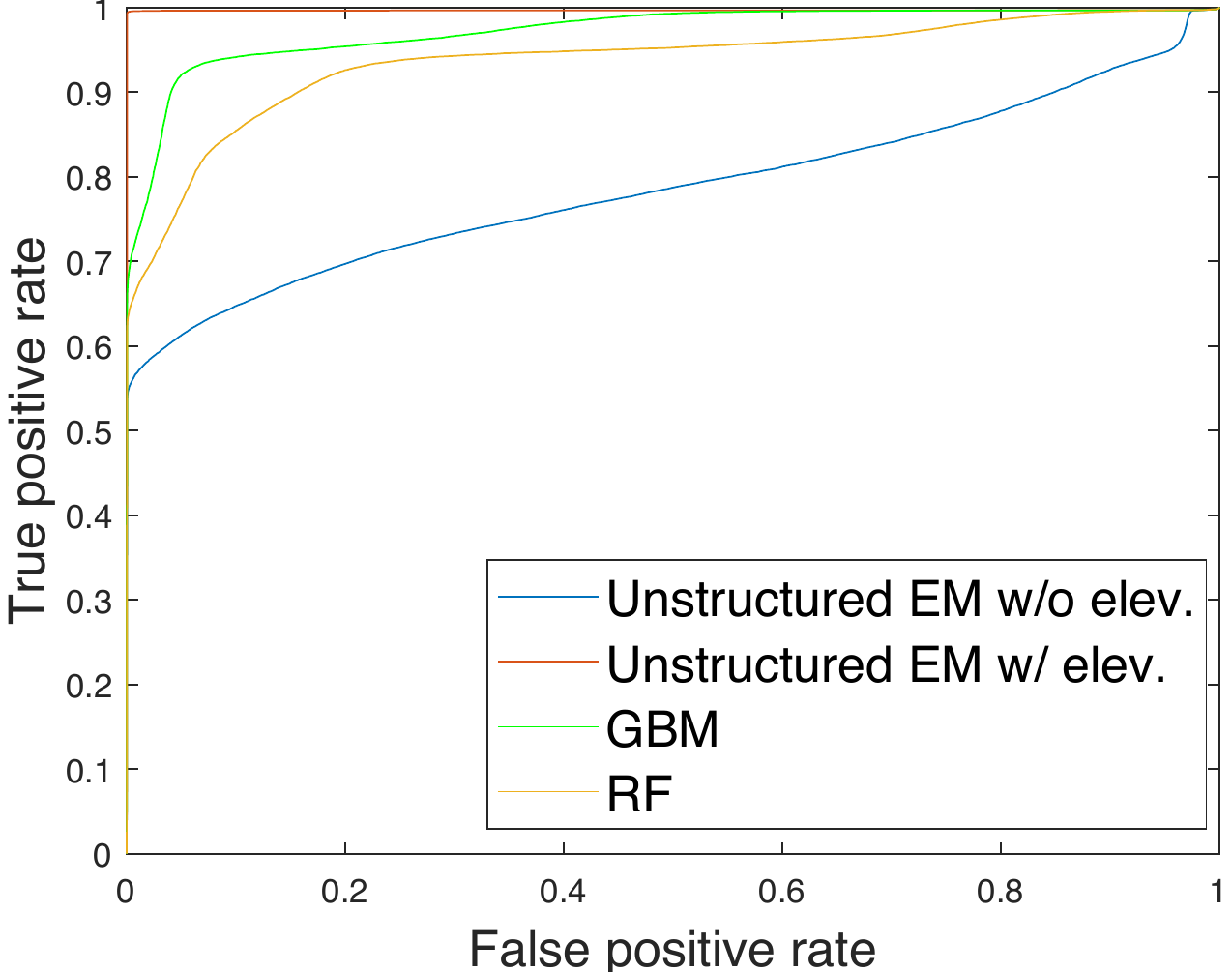}
    \caption{ROC curve on Harvey, Texas data}
    \label{fig:roc}
   \end{figure}

Moreover, we plot the ROC curve and calculate the AUC of ROC curve for unstructured EM, GBM and RF classifier. As shown in Figure~\ref{fig:roc} and Table~\ref{tab:aucroc}, the unstructured EM with elevation. gives the best ROC curve with AUC of 0.996, while the unstructured EM without elevation. shows the less significant result with AUC of 0.786. It means the spatial features play an important role in the unstructured EM classifier. The two baseline methods random forest and GBM show better results than unstructured EM without elevation classifier. 

   \begin{table}[h]\footnotesize
\centering
\caption{Comparison on AUC of ROC curve}
\label{tab:aucroc}
\begin{tabular}{cccccc}
\hline
Classifiers & AUC \\ \hline
{GBM}&{0.975}\\ \hline  
{RF}&{0.946}\\ \hline  
{Unstructured EM w/o elev.}&{0.786}\\  \hline
{Unstructured EM  w/ elev.}&{0.996}\\ \hline 
\end{tabular}
\end{table}

 
\subsection{Salt-and-pepper Noise}
\begin{table}[h]\footnotesize
\centering
\caption{Comparison on the total number of salt-and-pepper noise (The total number of pixels is 19,167,008)}
\label{tab:compgam}
\begin{tabular}{cccccc}
\hline
Classifiers & Number of Salt-and-pepper noise \\ \hline
{GBM}&{62,846}\\ \hline  
{RF}&{69,354}\\ \hline  
{Unstructured EM w/o elev.}&{77,625}\\  \hline
{Unstructured EM  w/ elev.}&{23,803}\\ \hline 
{EM Structured }&{3,740}\\ \hline 
\end{tabular}
\end{table}

We used a spatial autocorrelation statistic called Gamma index~\cite{anselin1995local} to quantify the salt-and-pepper noise level. 
Gamma index measures the similarity between the attribute values of a location and those of its neighbors. It is defined as 
\begin{equation}\label{eq:empostll}
    \begin{split}
    \Gamma_i=\frac{\sum_j W_{i,j} I_i I_j }{\sum_j W_{i,j} }
    \end{split}
\end{equation}
where i and j are locations, $W_{i,j}$ is 1 if j is i’s neighbor, and 0 otherwise,  $I_i$ is 1 if pixel i is flood pixel and -1 otherwise.  We define a salt-and-pepper noise pixel as a pixel that has a negative local Gamma index ($\Gamma_i<0$). We calculated the total number of salt-and-pepper noise pixels across all locations in the predicted class map. The total numbers of salt-and-pepper noise pixels of the five classifiers are summarized in Table~\ref{tab:compgam}.  We can see that RF, GBM, and EM i.i.d. without elevation have the highest salt-and-pepper noise level (above 60,000). EM i.i.d. with elevation is much better with only 23,803 salt-and-pepper noise. The EM structured method has the lowest number of salt-and-pepper noise, i.e., 3,740, about one order of magnitude lower than other classifiers.


\subsection{Effect of the ratio of labels}
\begin{figure}
    \vspace{-1mm}
    \centering
    \includegraphics[height=2in]{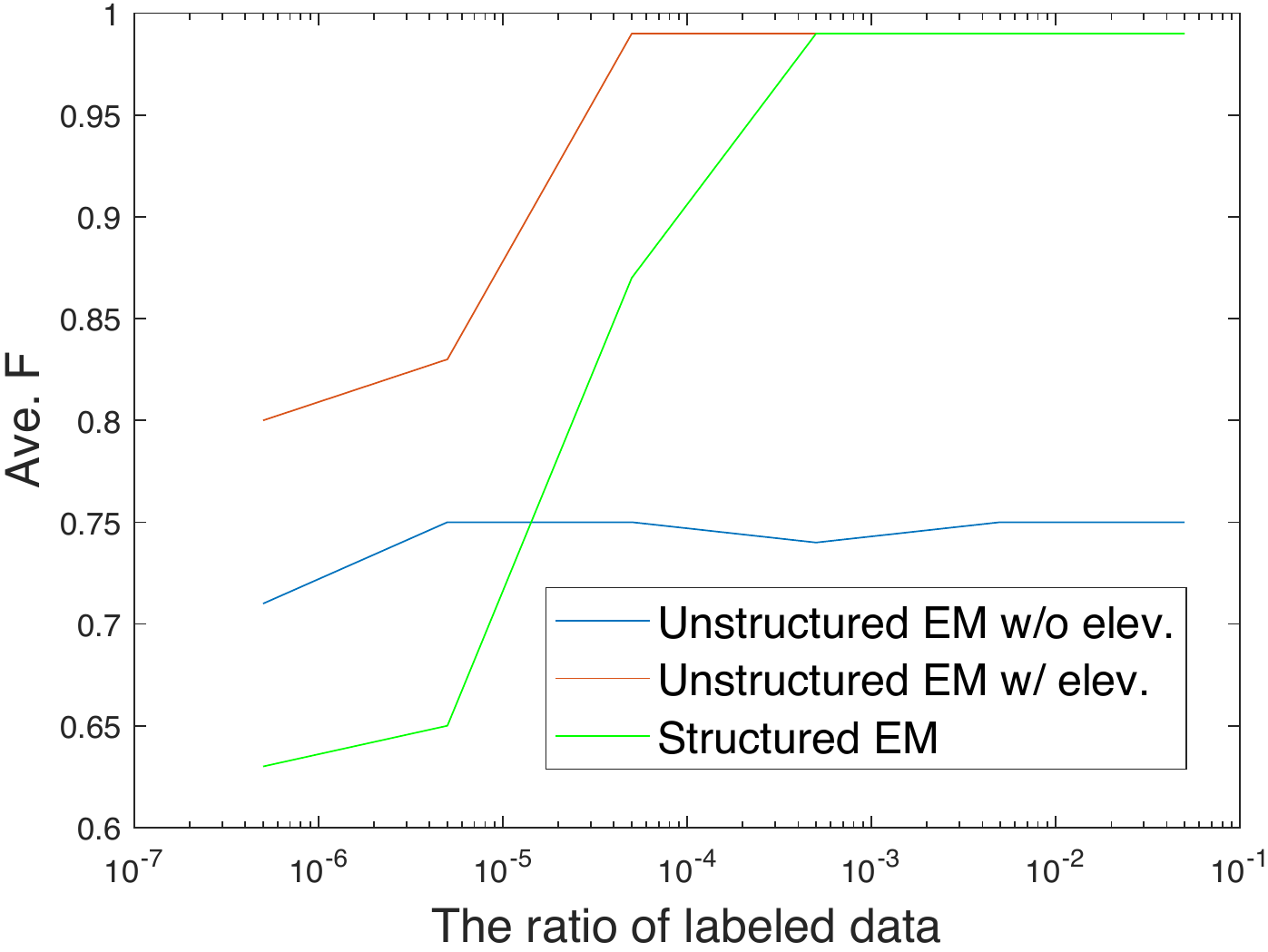}
    \caption{Sensitivity to the ratio of labeled samples}
    \label{fig:ratiooflabel}
   \end{figure}

We analyzed the sensitivity of three candidate EM methods to the ratio of training labels. Specifically, we increased the ratio of labeled samples from $5\times 10^{-7}$ to $5 \times 10^{-2}$. The results were summarized in Figure~\ref{fig:ratiooflabel}. We can see that as the ratio of labeled samples increase, the F-scores of all three methods first improved and then converged to an optimal value. Specifically, EM i.i.d. (represented by Gaussian mixture model) achieved the lowest peak F-score (around 0.75), while EM i.i.d. with elevation feature and EM structured model (represented by HMT) have a much better peak F-score (around 0.99). We also observe that when the ratio of labeled sample is small (e.g., below $10^{-5}$), EM structured model has a poor performance (with an F-score below 0.65). This is probably because more labeled samples are needed to initialize good parameters for our representative EM structured model (HMT). 

\section{Conclusions and Future work}
This paper makes a comparative study between unstructured and structured EM in semi-supervised learning. We compare the two methods in their theoretical properties and find that EM-based semi-supervised learning can be considered as a generalization of self-training method with soft class assignment on unlabeled samples. A case study on flood mapping datasets shows that unstructured EM method can be significantly impacted by noise and obstacles in sample features. Adding spatial contextual features in unstructured EM method can reduce the impact of noise and obstacles but will still produce salt-and-peper noise errors. Finally, structured EM can better address the issue compared with the other methods in this flood mapping applications. In future work, we plan to conduct the comparison studies on more types of datasets and applications to see if the conclusion can hold in general.

\section*{Acknowledgement}
This material is based upon work supported by the NSF under Grant No. IIS-1850546, IIS-2008973, CNS-1951974 and the University Corporation for Atmospheric Research (UCAR).




\bibliographystyle{IEEEtran}
\bibliography{ref}

\begin{IEEEbiography}[{\includegraphics[width=1in, height=1.25in, clip, keepaspectratio]{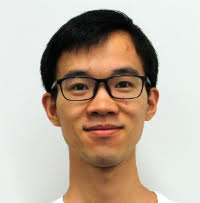}}]
{Wenchong He} is a Ph.D. student in the department of Computer Science at the University of Alabama. He received his B.S. degree from University of Science and Technology of China (USTC) in 2017, and Master degree from the College of William Mary in 2019. His research interests include machine learning, data mining and deep learning. 
\end{IEEEbiography}

\begin{IEEEbiography}[{\includegraphics[width=1in, height=1.25in, clip, keepaspectratio]{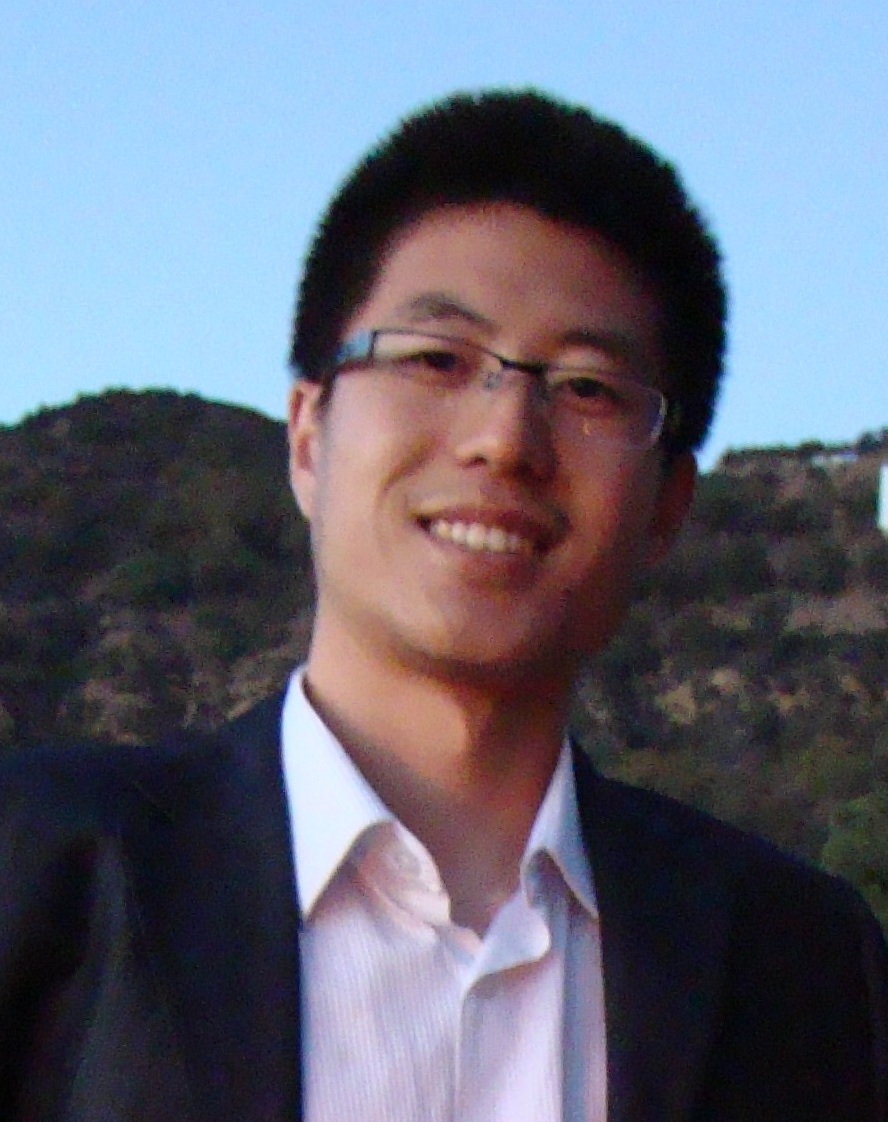}}]
{Zhe Jiang} is an assistant professor in the department of Computer Science at the University of Alabama. He received his Ph.D. from the University of Minnesota, Twin Cities in 2016, and B.E. from the the University of Science and Technology of China in 2010. His research interests include spatial and spatio-temporal data mining, spatial database, and geographic information system. He is a member of IEEE.
\end{IEEEbiography}





%




\end{document}